\pdfoutput=1
\documentclass[11pt]{article}

\usepackage[preprint]{acl}

\usepackage{times}
\usepackage{latexsym}

\usepackage[T1]{fontenc}
\usepackage[utf8]{inputenc}

\usepackage{microtype}

\usepackage{inconsolata}

\usepackage{graphicx}

\usepackage{tabularx}
\usepackage{multirow}
\usepackage{multicol}
\usepackage{url}
\usepackage{booktabs}
\usepackage{xcolor}
\usepackage{colortbl}
\usepackage{algorithm}
\usepackage{algorithmic}
\usepackage{tcolorbox}
\usepackage{pifont}
\usepackage{amsmath}
\usepackage{amsthm}
\usepackage{amssymb}
\usepackage{relsize}
\usepackage{longtable}
\usepackage{enumitem}
\usepackage{enumerate}
\usepackage{xspace}
\usepackage{subcaption}
\usepackage{fontawesome}
\usepackage{bm} % for bold math
\usepackage{siunitx}
\usepackage{nicefrac}       % compact symbols for 1/2, etc.
\usepackage{microtype}      % microtypography
\usepackage{colortbl}
\usepackage{tabularx}
\usepackage{makecell}
\usepackage{longtable}
\usepackage{float}
\usepackage{arydshln}
\usepackage{xcolor}
\usepackage{dutchcal}
\usepackage[capitalize,noabbrev]{cleveref}

\definecolor{lightgreen}{rgb}{0.88, 1, 0.88}
\definecolor{lightpurple}{RGB}{180, 150, 255}
\definecolor{lightred}{RGB}{255, 204, 203}
\definecolor{lightblue}{RGB}{173, 216, 230}
\usepackage{xcolor}
\usepackage{tcolorbox}
\tcbuselibrary{skins}
\newtcolorbox{promptbox}[1][]{
  enhanced,
  colback=gray!8,         % 背景
  colframe=black,         % 边框色
  boxrule=0.6pt,          % 边框线宽
  arc=3mm,                % 圆角
  left=3mm,right=3mm,top=2mm,bottom=2mm, % 内边距
  drop shadow=black!30,   % 阴影
  title=#1,               % 标题文字
  fonttitle=\bfseries,    % 标题加粗
}

\newcommand{\eg}{e.g.\xspace}

\newcommand{\method}{\ensuremath{\textnormal{\textsc{MemMA}}}\xspace}
\newcommand{\nop}[1]{}

\title{MemMA: Coordinating the Memory Cycle through Multi-Agent Reasoning and In-Situ Self-Evolution}

\author{Minhua Lin$^{1}$, Zhiwei Zhang$^{1}$, Hanqing Lu$^2$, Hui Liu$^3$,\\ \textbf{Xianfeng Tang}$^{3}$, \textbf{Qi He}$^3$, \textbf{Xiang Zhang}$^1$, \textbf{Suhang Wang}$^1$ \\  
 $^{1}$The Pennsylvania State University $^{2}$Amazon $^{3}$Microsoft\\
\texttt{\{mfl5681,szw494\}@psu.edu}\\ 
}

\begin{document}
\maketitle

\begin{abstract}
Memory-augmented LLM agents maintain external memory banks to support long-horizon interaction, yet most existing systems treat construction, retrieval, and utilization as isolated subroutines. This creates two coupled challenges: \emph{strategic blindness} on the forward path of the memory cycle, where construction and retrieval are driven by local heuristics rather than explicit strategic reasoning, and \emph{sparse, delayed supervision} on the backward path, where downstream failures rarely translate into direct repairs of the memory bank. To address these challenges, we propose \method, a plug-and-play multi-agent framework that coordinates the memory cycle along both the forward and backward paths. On the forward path, a Meta-Thinker produces structured guidance that steers a Memory Manager during construction and directs a Query Reasoner during iterative retrieval. On the backward path, \method introduces in-situ self-evolving memory construction, which synthesizes probe QA pairs, verifies the current memory, and converts failures into repair actions before the memory is finalized. Extensive experiments on LoCoMo show that \method consistently outperforms existing baselines across multiple LLM backbones and improves three different storage backends in a plug-and-play manner. Our code is publicly available at 
% \url{https://anonymous.4open.science/r/memma-588C}.
\url{https://github.com/ventr1c/memma}.
\end{abstract}
% \begin{abstract}
% Memory-augmented LLM agents maintain external memory banks to support long-horizon interaction, yet most treat construction, retrieval, and utilization as isolated subroutines.
% We identify \emph{strategic blindness} as a key failure mode: agents possess the mechanisms to edit and query memory but lack the meta-cognition to coordinate these operations across the full memory cycle.
% To address this, we propose \method, a multi-agent framework that can be plugged into existing memory systems and leverages the memory cycle effect to coordinate memory operations along two paths.
% First, a \emph{forward coordination path} separates strategic reasoning from low-level execution: a Meta-Thinker produces structured guidance that steers a Memory Manager during construction and diagnoses information gaps to direct a Query Reasoner during iterative retrieval.
% Second, a \emph{backward self-evolution path} injects dense intermediate feedback into construction by synthesizing probe QA pairs, verifying the memory against them, and repairing detected omissions before the memory is committed.
% On the LoCoMo benchmark, \method improves the strongest baseline by +5.92 LLM-as-a-Judge accuracy, with the largest gains on multi-hop and temporal questions, and consistently improves three different storage backends across LLM backbones. Our code is publicly available at \url{https://anonymous.4open.science/r/memma-588C}.
% \end{abstract}

\section{Introduction}
Large language models (LLMs)~\cite{radford2018improving,radford2019language,touvron2023llama} are evolving from episodic chatbots into persistent \emph{agentic} systems~\cite{wang2024survey,yao2022react,yang2024swe} that execute complex workflows over days or weeks.
In such settings, agents receive a continuous stream of observations---user constraints, tool outputs, and environmental feedback---whose consequences unfold over long horizons.
This shift makes \emph{controllable, long-term memory} a first-class requirement: relying solely on ephemeral context windows is insufficient, as they are computationally expensive and prone to attention dilution.
To maintain coherence over time, agents must actively manage an external memory bank~\cite{packer2023memgpt,hu2025memory}, deciding what to retain and how to retrieve it under uncertainty.

\begin{figure}[t]
    \small
    \centering
        \includegraphics[width=0.95\linewidth]{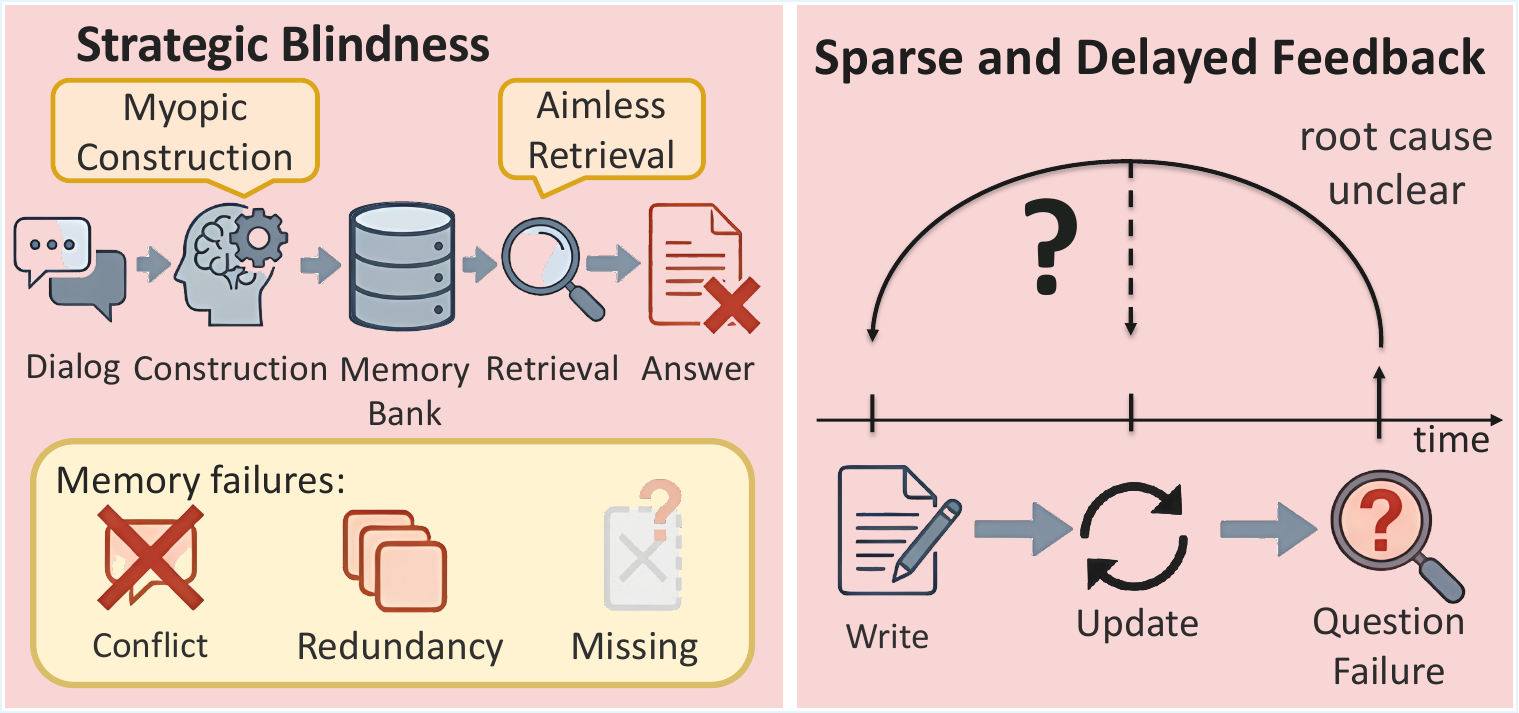}
        % \vskip -1em
    % \vskip -1.em
    \caption{Two challenges in leveraging the memory cycle effect.}
    \vspace{-1.2em}
    \label{fig:challenge_memory_cycle}
\end{figure}

Effective memory, however, is not merely a storage utility; it is a closed-loop dynamic, conceptualized as the \textbf{memory cycle effect}~\cite{zhang2025learn}.
This cycle has three coupled phases: \emph{construction}, \emph{retrieval}, and \emph{utilization}.
Construction determines what information enters the memory bank and how it is organized; retrieval determines what stored information is surfaced as evidence; and utilization reveals whether the retrieved evidence is sufficient for downstream reasoning.
This coupling implies that optimizing these stages in isolation is fundamentally suboptimal: a retrieval failure may stem from a much earlier construction error, while utilization outcomes should ideally feed back to improve future memory decisions.
Despite this intrinsic dependency, most existing memory-augmented agents~\cite{chhikara2025mem0,fang2025lightmem,xu2025mem,yan2025memory,zhou2025mem1,shen2026membuilder} still treat memory operations as isolated, reactive subroutines, overlooking the coupling between stages.
To leverage the memory cycle effect, two technical challenges must be addressed (Fig.~\ref{fig:challenge_memory_cycle}).

First, on the \emph{forward path} of the memory cycle, current systems often suffer from \textbf{strategic blindness}: they possess the mechanisms to edit memory and issue retrieval queries, yet lack explicit \textit{meta-cognition to coordinate these actions toward downstream question answering}.
As our preliminary analysis shows (Sec.~\ref{ssec:motivate_study}), this manifests as two pathologies:
(i)~\emph{Myopic Construction}, where the agent accumulates or overwrites conflicting facts without resolution; and
(ii)~\emph{Aimless Retrieval}, where the agent performs shallow or repetitive searches without narrowing the true information gap.
These failures suggest that \textit{effective forward-path memory behavior requires explicit coordination between construction and retrieval, rather than isolated, short-sighted decisions}.
% These failures suggest that construction and retrieval should not be driven by isolated, short-sighted decisions, but instead require explicit coordination along the forward path.
% As a result, memory construction and retrieval are often driven by local heuristics rather than globally coherent reasoning.

Second, on the \emph{backward path} of the memory cycle, feedback from utilization to construction is typically \emph{sparse and delayed}.
Whether a memory-writing decision is useful may become clear only much later, when the agent fails a downstream question.
This makes credit assignment difficult: when an answer is wrong, it is hard to identify which earlier construction decision caused the failure, allowing omissions and unresolved conflicts to persist in the memory bank and affect later updates.
Although recent methods use reflection or experiential learning to improve agent behavior~\cite{shinn2023reflexion,zhao2024expel,zhang2026memrl}, downstream failures are still rarely converted into direct signals for repairing the memory bank itself.

To address these challenges, we propose \method\ (\textbf{Mem}ory Cycle 
\textbf{M}ulti-\textbf{A}gent Coordination), a plug-and-play multi-agent 
framework that coordinates the memory cycle along its forward and backward paths.
Specifically, for the \emph{forward path}, \method separates strategic reasoning from low-level execution through a planner--worker architecture: a Meta-Thinker produces structured guidance that steers a Memory Manager during construction (what to retain, consolidate, or resolve), {thereby mitigating \emph{Myopic Construction}}, and directs a Query Reasoner during retrieval {by diagnosing missing evidence and how to retrieve it, replacing one-shot search with diagnosis-guided iterative refinement and thereby mitigating \emph{Aimless Retrieval}}. 
For the \emph{backward path}, \method introduces {in-situ self-evolving memory construction}: after each session, the system synthesizes probe QA pairs, verifies the memory against them, and converts failures into repair actions on the memory bank through evidence-grounded critique and semantic consolidation, 
before the memory is committed for future use. 
{This directly addresses \emph{sparse and delayed supervision} by turning 
downstream failures into immediate, localized repair signals for the current 
memory state, before flawed memories propagate into future memory updates.}
% Together, these two mechanisms enable \method to coordinate the memory cycle along both its forward and backward paths.

Our contributions are:
(i) \textbf{Analysis}. We identify two technical challenges in leveraging the memory cycle effect: \emph{strategic blindness} on the forward path and \emph{sparse, delayed feedback} on the backward path, and provide empirical evidence through a controlled preliminary study (Sec.~\ref{ssec:motivate_study}).
(ii) \textbf{Framework}. We propose \method, a plug-and-play multi-agent framework that coordinates the memory cycle along both its forward and backward paths, combining reasoning-aware coordination for construction and iterative retrieval with in-situ self-evolving memory construction for backward repair.
(iii) \textbf{Experiments}. \method outperforms existing baselines on LoCoMo across multiple LLM backbones, and consistently improves three storage backends as a plug-and-play module.

\section{Related Work}
\noindent\textbf{Memory-Augmented LLM Agents.}
External memory has become a core component of LLM agents that operate over long horizons.
Prior work improves long-term memory from several directions, including memory architecture~\cite{packer2023memgpt,zhong2024memorybank}, memory organization and consolidation~\cite{xu2025mem,fang2025lightmem}, and memory retrieval~\cite{du2025memr}.
These methods substantially improve individual stages of the memory pipeline, but they primarily optimize storage, organization, or retrieval in isolation.
Our work is inherently different from existing work: \method jointly coordinates memory construction and iterative retrieval, and converts utilization failures into direct repair signals for the memory bank.
% By contrast, \method addresses a broader scope: it coordinates both construction and retrieval along the forward path of the memory cycle, and further converts utilization failures into direct repair signals for the memory bank along the backward path. 
Full version is in Appendix~\ref{appendix:full_related_works}.

% \noindent\textbf{Self-Evolution for LLM Agents.}
% Recent work improves agent behavior through output-level 
% refinement~\cite{madaan2023self}, 
% experience accumulation~\cite{zhao2024expel,wang2023voyager}, 
% and policy optimization for memory~\cite{shen2026membuilder,yan2025memory}.
% These approaches improve responses, maintain separate experience 
% stores, or optimize memory policies through training, but do not 
% directly repair the memory bank during construction.
% \method fills this gap by verifying the current memory with probe 
% QA pairs and converting failures into repair actions before the 
% memory is committed. Full version is in Appendix~\ref{appendix:self_evolve_LLM_related_works}.
\section{Preliminaries and Motivation}
\subsection{Problem Setting}
\noindent\textbf{Task Setup.}
We consider a long-horizon conversational setting in which an agent processes a stream of dialogue chunks
$\mathcal{C}=\{c_1,\ldots,c_T\}$ over time. 
The stream is further organized into sessions $\mathcal{S}=\{s_1,\ldots,s_N\}$,where each session $s_\tau$ consists of one or more consecutive chunks corresponding to a coherent interaction episode.
At each step $t$, the agent maintains an external memory bank $M_t$ composed of structured entries
(e.g., text, timestamp, source, and speaker metadata), which is updated as new conversational information arrives.
After processing the full stream $\mathcal{C}$, the agent is evaluated on a set of questions $Q$.
For each query $q \in Q$, it retrieves evidence $E(q)$ from $M_T$ and outputs an answer $\hat{y}(q)$.
Our goal is to design an agent $\pi$ that maximizes answer accuracy by jointly improving memory construction and retrieval.

% After observing the full chunks $\mathcal{C}$, the agent is evaluated on a set of queries ${Q}$: for each $q\in{Q}$, it retrieves evidence $E(q)$ from $M_T$ and outputs an answer ${y}(q)$. Our objective is to learn $\pi$ that maximizes answer accuracy by jointly optimizing memory construction and retrieval. 

\noindent\textbf{Challenges.}
This setting is challenging because success depends on both memory construction and memory retrieval.
During \emph{construction}, the agent must decide what to write, update, merge, or discard when a new chunk arrives.
During \emph{retrieval and answering}, it must identify the right evidence from memory under ambiguity, temporal dependencies, and incomplete or underspecified initial queries.
The challenge is therefore not merely to improve answer generation, but to maintain a useful memory bank and retrieve the right evidence under bounded memory and retrieval budgets.

\subsection{Memory Cycle Effect as a Design Lens}
\label{ssec:cycle_effect}
The above challenges suggest that long-term memory should not be viewed as a linear pipeline of isolated modules.
Instead, we adopt the \textbf{memory cycle effect}~\cite{zhang2025learn} as a design lens for analyzing long-term memory systems.
Under this view, memory forms a closed loop with three tightly coupled phases: \emph{construction}, \emph{retrieval}, and \emph{utilization}.
Construction determines what information enters the memory bank and how it is organized; retrieval determines what stored information is surfaced as evidence; and utilization reveals whether the retrieved evidence is sufficient for downstream answering.

This perspective highlights two dependencies. First, there is a \emph{forward dependency}: construction constrains retrieval, and retrieval in turn constrains utilization.
A poorly constructed memory bank may omit important details, retain redundant entries, or leave conflicts unresolved, all of which degrade downstream retrieval quality. 
Second, there is a \emph{backward dependency}: utilization outcomes expose deficiencies in upstream memory operations, since answering failures may stem from earlier storage omissions, unresolved contradictions, or poorly targeted retrieval. 
As a result, the utility of memory operations is often sparse and delayed, making isolated optimization of memory modules fundamentally suboptimal.

Together, these dependencies suggest that long-term memory should be studied as a coupled cycle rather than independent storage and retrieval components.
This motivates the need for mechanisms that explicitly coordinate forward memory execution and propagate utilization feedback backward to improve future memory decisions.

\begin{table}[t]
\centering
% \scriptsize
% \small

% \vspace{-1em}
\caption{Preliminary analysis results (\%) in LoCoMo dataset, GPT-4o-mini is the backbone LLM. 
% \zhiwei{since the performance gap for F1 is not that significant, could we report variance}
} 
% \vspace{-1em}
\label{tab:prior_study}
\scalebox{0.95}{\begin{tabular}{lccc}
\toprule
\textbf{Method} & \textbf{F1} & \textbf{B1} & \textbf{ACC} \\%& 
     \midrule
Static Baseline & 22.64 & 17.24 & 52.60 \\
Unguided Active & 23.49 & \textbf{18.36} & 54.60\\
Strategic Active & \textbf{24.78} & 17.73 & \textbf{59.21}\\
\bottomrule
\end{tabular}}
% \vspace{-1.5em}
\end{table}

\subsection{Motivating Analysis: Strategic Blindness}
\label{ssec:motivate_study}
% \suhang{the logic here is quite strange. We were talking about ``the need for xxx'' in the above paragraph. Then we suddenly talk about ``recent agents have moved beyond fully passive to xxx''. I do not get the connection or the reason why we want to discuss this. Are we trying to do experiments to further demonstrate ``the need for mechanisms that xxx'', or are we trying to show another issue? Could we add some transition sentences here?} 
The analysis above motivates coordination across the memory cycle, 
but do existing active memory agents achieve this in practice?
Recent agents~\cite{fang2025lightmem,xu2025mem} have moved beyond 
fully passive memory by introducing active updates or iterative 
retrieval. However, most still operate in a largely reactive 
manner: they trigger operations based on local context or immediate similarity signals rather than an explicit global strategy. 
% Recent memory-augmented agents~\cite{fang2025lightmem,xu2025mem} have moved beyond fully passive memory by introducing active updates or iterative retrieval.
% However, most of them still operate in a largely reactive manner: they trigger operations based on local context or immediate similarity signals rather than an explicit global strategy. 
We characterize this limitation as \emph{strategic blindness}: the agent has the \emph{hands} to edit memory and issue retrieval queries, but lacks the \emph{brain} to coordinate these actions across the full memory cycle.
% We identify two concrete manifestations:
This manifests as:
(i)~\emph{Myopic Construction}: construction decisions are driven 
by local context rather than downstream utility. The agent 
indiscriminately appends, overwrites, or ignores information, 
leaving redundancy and conflicts unresolved.
(ii)~\emph{Aimless Retrieval}: when the initial query is incomplete 
or semantically mismatched with stored memory, one-shot retrieval 
or shallow rewrites fail to surface the required evidence. Without 
strategic guidance, successive queries do not narrow the information gap.

\noindent\textbf{Setup.} 
To empirically validate this diagnosis, we conduct a preliminary study on a subset of LoCoMo~\cite{maharana2024evaluating}, focusing on reasoning-intensive queries by excluding adversarial samples.
We compare three progressively stronger baselines using GPT-4o-mini~\cite{gpt4ocard} as the backbone:
(i) \textit{Static}, which performs memory construction followed by one-shot top-$30$ retrieval;
(ii) \textit{Unguided Active}, which adds iterative query rewriting without strategic guidance; and
(iii) \textit{Strategic Active}, which introduces a planner to guide both construction and retrieval.
We report token-level F1, BLEU-1 (B1), and LLM-as-a-Judge accuracy (ACC).
More evaluation details are provided in Appendix~\ref{appendix:preliminary_evaluation_details}.

% \noindent\textbf{Empirical analysis.} 
% Table~\ref{tab:prior_study} reveals two findings:
% (i)~\emph{Refinement provides capability:} the \textit{Unguided Active} variant (54.6\% ACC) outperforms the \textit{Static} variant (52.6\%), confirming that iterative refinement helps bridge the semantic gap missed by one-shot retrieval;
% (ii)~\emph{Reasoning provides control:} the \textit{Strategic Active} variant achieves a larger leap to 59.2\% ACC, substantially outperforming both weaker variants.
% Case studies in Appendix~\ref{appendix:preliminary_case_studies} further show that unguided memory storage and retrieval often induce reasoning drift and degrade utility.
% These findings suggest that active memory operations alone are insufficient: explicit strategic reasoning is needed to guide both construction and retrieval. \suhang{how does this experiment verify Myopic
% Construction and Aimless Retrieval? Could you explicitly connect the experiment design with these two points?}

\noindent\textbf{Empirical analysis.} Table~\ref{tab:prior_study} reveals two findings: (i)~\emph{Refinement provides capability:} {Unguided Active} (54.6\% Acc) outperforms {Static} (52.6\%), confirming that one-shot retrieval often fails to surface the required evidence when the initial query is incomplete or mismatched with memory, which directly reflects \emph{Aimless Retrieval}. (ii)~\emph{Reasoning provides control:} {Strategic Active} achieves a larger leap to 59.2\% Acc. Since it shares the same active operators as {Unguided Active}, this gap reflects the value of explicit strategic guidance in addressing both \emph{Aimless Retrieval} and \emph{Myopic Construction}. Case studies in Appendix~\ref{appendix:preliminary_case_studies} further illustrate both pathologies with concrete examples of redundant entries and retrieval drift. 
These findings suggest that active memory operations alone are insufficient: explicit strategic reasoning is needed to guide both construction and retrieval. 

\section{Methodology}
\label{sec:methodology}
\begin{figure}[t]
    \small
    \centering
        \includegraphics[width=0.95\linewidth]{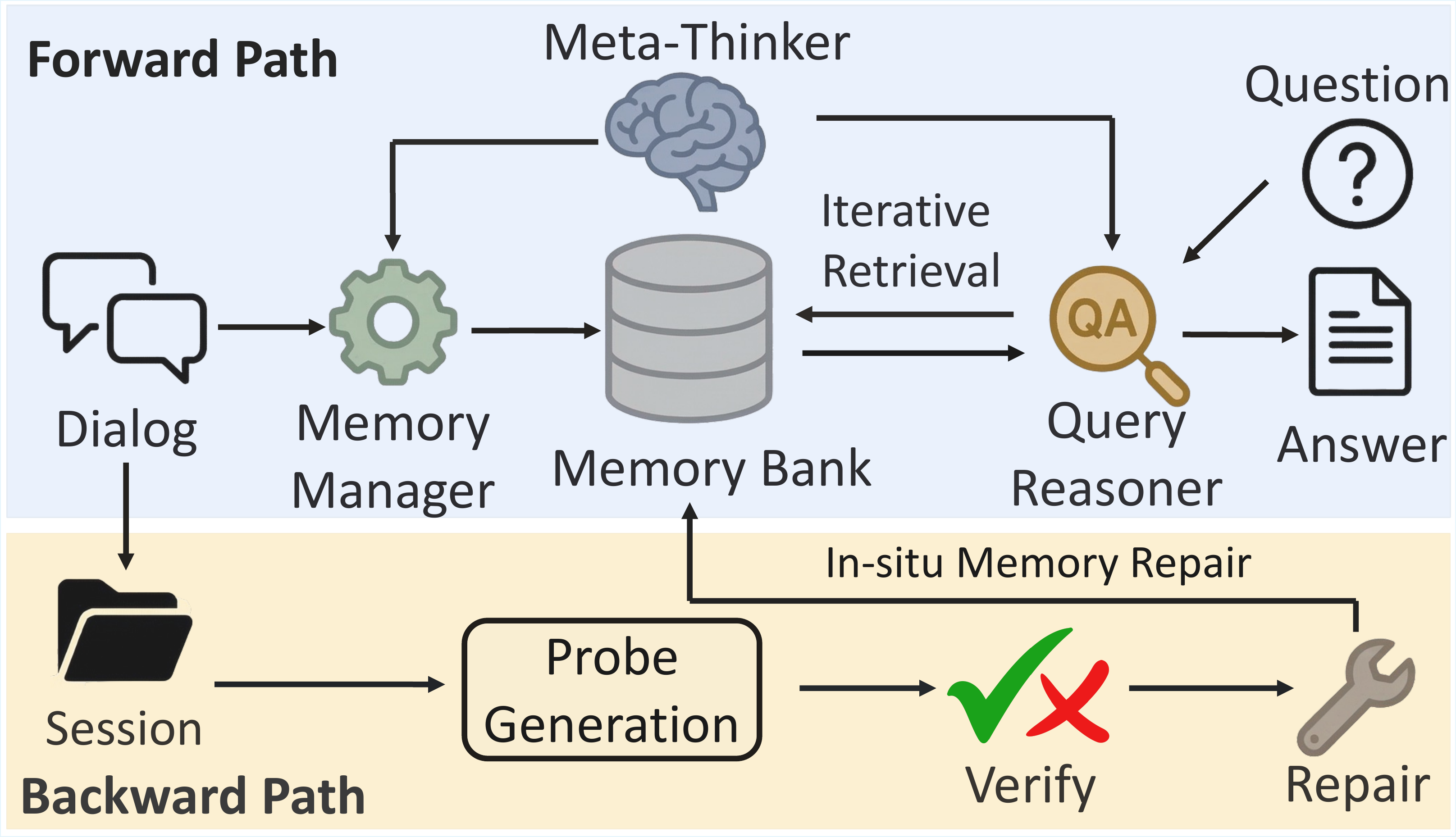}
        % \vskip -1em
    % \vskip -0.5em
    \caption{Overview of \method.}
    \vspace{-1.2em}
    \label{fig:framework_memma}
\end{figure}

% We present \method, a plug-and-play multi-agent framework for coordinating the memory cycle.
% The framework has two parts.
% Sec.~\ref{ssec:architecture_design} describes the forward path, which handles online memory construction and iterative retrieval through a planner--worker architecture.
% Sec.~\ref{ssec:self_evolving_construction} describes the backward path, which uses in-situ self-evolution to repair the current memory before it is committed. 
% \suhang{I think we should rewrite this paragraph.  Directly start with something like ``Motivated by the xxx, we propose xxx'', then give the high-level idea of the proposed framework. This will have a smoother logic}
Motivated by the memory cycle effect (Sec.~\ref{ssec:cycle_effect}) and strategic blindness (Sec.~\ref{ssec:motivate_study}), we present \method, a plug-and-play multi-agent framework that coordinates the memory cycle along its forward and backward paths (Fig.~\ref{fig:framework_memma}). Sec.~\ref{ssec:architecture_design} describes the forward path: a planner--worker architecture that separates strategic reasoning from low-level execution to address strategic blindness. 
Sec.~\ref{ssec:self_evolving_construction} describes the backward path: an in-situ self-evolution mechanism that addresses sparse, delayed feedback by generating synthetic probe QA immediately after each session, providing dense, localized supervision for memory repair before the current memory is committed. 
% Sec.~\ref{ssec:self_evolving_construction} describes the backward path: an in-situ self-evolution mechanism that replaces sparse, delayed utilization feedback with dense, immediate supervision through synthetic probe QA, enabling memory repair before the current memory is committed. 
% The overall framework is illustrated in Fig.~\ref{fig:framework_memma}.

\subsection{Reasoning-Aware Coordination over the Forward Path}
\label{ssec:architecture_design}
% Motivated by the memory cycle effect (Sec.~\ref{ssec:cycle_effect}) and strategic blindness (Sec.~\ref{ssec:motivate_study}), 
\method coordinates online construction, iterative retrieval, and answer-time utilization through specialized yet tightly coupled agents.
Its key design principle is to separate strategic reasoning (what to store, what is missing, and when to stop) from low-level execution (memory editing, evidence retrieval, and answer generation).

% We first describe the forward path of \method.
% Its key design principle is to separate strategic reasoning (what to store, what is missing, and when to stop) from low-level execution (memory editing, evidence retrieval, and answer generation).
% \method coordinates online construction, iterative retrieval, and answer-time utilization through specialized yet tightly coupled agents.

\noindent\textbf{Pipeline Overview.}
\method uses a planner--worker architecture with four roles:
(i)~a \textit{Meta-Thinker} $\pi_{p}$ for high-level strategic reasoning,
(ii)~a \textit{Memory Manager} $\pi_{s}$ for memory editing,
(iii)~a \textit{Query Reasoner} $\pi_{r}$ for iterative query refinement, and
(iv)~an \textit{Answer Agent} $\pi_{a}$ for final response generation.

During \emph{construction}, when a new dialogue chunk $c_t$ arrives, $\pi_{p}$ analyzes it against existing memory $M_{t-1}$ and produces meta-guidance on what to retain, consolidate, or resolve.
Conditioned on the guidance, $\pi_{s}$ selects an atomic edit to update $M_{t-1}$ to $M_t$.
During \emph{question answering}, given a query $q$, $\pi_{r}$ retrieves candidate evidence from $M_T$ and iteratively refines its search.
At each step, $\pi_{p}$ judges whether the current evidence is sufficient; if not, it identifies the most critical gap and directs $\pi_{r}$ to refine the query toward complementary evidence.
The loop ends when $\pi_{p}$ deems the evidence sufficient or a budget $H$ is reached. Then $\pi_{a}$ generates the final answer.
We detail each component below.

\noindent\textbf{Meta-Thinker $\pi_{p}$.}
$\pi_{p}$ is the planning layer of \method, responsible for both construction and retrieval guidance.
It produces phase-specific guidance conditioned on the current input and a bounded memory view:
\begin{equation}
% \small
\begin{aligned}
g_t^{S} &\sim \pi_{p}(\cdot \mid c_t,\tilde{M}_{t-1}), \\
% g_{q,h}^{R} &\sim \pi_{p}(\cdot \mid q,E_h,\tilde{M}_{T}),\\
g_{q,h}^{R} &\sim \pi_{p}(\cdot \mid q, E_h, U_h, \tilde{M}_{T}),
\end{aligned}
\label{eq:meta_guidance}
\end{equation}
where $g_t^{S}$ is construction guidance at step $t$ and $g_{q,h}^{R}$ is retrieval guidance at refinement step $h$.
Here, {$E_h$ denotes the evidence accumulated up to step $h$}, $U_h=\{u_0,\ldots,u_h\}$ denotes the query history, and $\tilde{M}$ denotes a \emph{bounded} view of the memory bank, e.g., top-$k$ recent or semantically related entries.
% Each guidance signal is a structured textual directive for the downstream worker, with phase-specific content.

% where $g_t^{S}$ is construction guidance at step $t$ and $g_{q,h}^{R}$ is retrieval guidance at refinement step $h$.
% Here, $U_h=\{u_0,\ldots,u_h\}$ denotes the query history, and $\tilde{M}$ denotes a \emph{bounded} view of the memory bank, such as top-$k$ recent and/or semantically related entries that fit within the context window. \suhang{what is $E_h$??}
% Each guidance signal is a structured textual directive for the downstream worker, with phase-specific content.

\emph{Construction.}
$g_t^{S}$ provides a set of \emph{focus points} that flag information importance, redundancy with existing entries, and potential conflicts.
These focus points steer $\pi_s$ toward globally consistent memories rather than indiscriminate accumulation.

\emph{Retrieval.}
$g_{q,h}^{R}$ is a critique of the current evidence $E_h$.
$\pi_{p}$ evaluates coverage, consistency, and specificity with respect to $q$.
If the evidence is sufficient, it returns \textsc{answerable}; otherwise, it returns \textsc{not-answerable} together with a diagnosis of what is missing and how to retrieve it, \eg, a missing attribute or temporal scope.
This encourages orthogonal evidence acquisition rather than near-duplicate searches.
Full guidance templates and examples are in Appendix~\ref{appendix:meta_thinker_details}.

\noindent\textbf{Memory Manager $\pi_{s}$.}
$\pi_{s}$ performs atomic memory edits based on the current chunk, bounded context, and guidance from $\pi_{p}$.
Given $c_t$, $\tilde{M}_{t-1}$, and $g_t^{S}$, it selects an action $a_t^{S} \in \{\texttt{ADD}, \texttt{UPDATE}, \texttt{DELETE}, \texttt{NONE}\}$:
\begin{equation}
\begin{aligned}
a_t^{S} &\sim \pi_{s}(\cdot \mid c_t,\tilde{M}_{t-1},g_t^{S}),\\
M_t &= \textsc{Apply}(M_{t-1}, a_t^{S}),
\end{aligned}
\label{eq:memory_manager}
\end{equation}
The guidance signal $g_t^{S}$ helps $\pi_{s}$ filter noise, consolidate redundancy, and resolve conflicts at the source rather than blindly appending.
$\pi_{s}$ is backend-agnostic and can wrap diverse memory implementations such as LightMem~\cite{fang2025lightmem} and A-Mem~\cite{xu2025mem}.

\noindent\textbf{Query Reasoner $\pi_r$.}
$\pi_r$ implements the \emph{active retrieval policy}.
To overcome the \emph{Aimless Retrieval} (Sec.~\ref{ssec:motivate_study}),
it replaces one-shot search with an iterative \emph{Refine-and-Probe} loop.
Let $u_0=q$ be the initial query and $U_h=\{u_0,\ldots,u_h\}$ the query history.
At step $h$, when $\pi_p$ deems the current evidence $E_h$ \textsc{not-answerable}, it emits guidance $g_{q,h}^{R}$.
$\pi_r$ then proposes the next query and retrieves additional evidence:
\begin{equation}
\begin{aligned}
u_{h+1} &\sim \pi_r(\cdot \mid U_h, E_h, g_{q,h}^{R}),\\
E_{h+1} &= E_h \cup \textsc{Search}(M_T, u_{h+1}).
\end{aligned}
\label{eq:query_reasoner}
\end{equation}
The loop terminates when $\pi_p$ returns \textsc{answerable} or the budget $H$ is reached.
Each refinement step targets the specific information gap diagnosed by $\pi_p$, so successive queries narrow the deficit rather than drifting across redundant rewrites. Full query rewrite prompt templates are in Appendix~\ref{appendix:query_reasoner_details}.

\noindent\textbf{Answer Agent $\pi_a$.}
Once the retrieval loop terminates, $\pi_a$ generates the final answer $\hat{y}(q)$ based on the query and the final evidence set $E(q)=E_H$:
\begin{equation}
\hat{y}(q) = F_{\pi_a}(q, E(q)),
\end{equation}
where $F_{\pi_a}$ denotes a generation function (\eg, an LLM call).
In our experiments, $\pi_a$ is kept frozen to decouple answer-generation capacity from memory quality, so that gains can be attributed to coordination over the memory cycle rather than to the parametric knowledge of $\pi_a$.

\subsection{In-Situ Self-Evolving Memory Construction}
\label{ssec:self_evolving_construction}
A major bottleneck in the memory cycle is that feedback for construction is typically sparse and delayed.
The utility of a storage decision made in session $\tau$ may become observable only much later, when the agent fails a downstream question.
Optimizing construction solely from final-task outcomes makes credit assignment difficult and lets early omissions propagate uncorrected. To address this, we introduce \textbf{in-situ self-evolving memory construction}, which provides dense intermediate feedback for the construction stage.
Instead of waiting for a future user query to expose a memory failure, \method synthesizes a set of probe QA pairs after each session and uses them to verify and repair the current memory before it is committed.

\noindent\textbf{Probe Generation.}
Let $s_\tau$ denote the current session, and let $M_{\tau}^{(0)}$ denote the provisional memory state obtained after applying the construction policy of Sec.~\ref{ssec:architecture_design} to $s_\tau$.
To obtain intermediate supervision, we construct a probe set
\begin{equation}
\mathcal{Q}_\tau = \{(q_j, y_j)\}_{j=1}^{J},
\end{equation}
where each $(q_j, y_j)$ is a synthetic question--answer pair grounded in $s_\tau$ and its relevant historical context $\tilde{M}_{\tau-1}$.
The questions are designed to test whether the provisional memory faithfully captures and can retrieve information introduced in the current session, covering single-session factual recall, cross-session relational reasoning, and temporal inference~\cite{shen2026membuilder}.
This turns a delayed end-task signal into $J$ localized supervision signals immediately after construction.
% In our setting, probe generation is triggered after each session.
Design details are in Appendix~\ref{appendix:synthetic_qa_details}.

% Let $s_\tau$ denote the current session, and let $M_{\tau}^{(0)}$ denote the provisional memory state obtained after applying the construction policy of Sec.~\ref{ssec:architecture_design} to $s_\tau$.
% To obtain intermediate supervision, we construct a probe set
% \begin{equation}
% \mathcal{Q}_\tau = \{(q_j, y_j)\}_{j=1}^{J},
% \end{equation}
% where each $(q_j, y_j)$ is a synthetic question--answer pair grounded in $s_\tau$ and its relevant historical context $\tilde{M}_{\tau-1}$.
% The questions are designed to test whether the memory bank correctly captured and can retrieve information from the current session, covering single-session factual recall, cross-session relational reasoning, and temporal inference~\cite{shen2026membuilder}.
% This converts one delayed end-task signal into $J$ localized supervision signals immediately after construction.
% In practice, probe generation can be triggered after every session or after a batch of sessions.
% Design details are in Appendix~\ref{appendix:synthetic_qa_details}.

\noindent\textbf{In-situ Verification.}
Given $\mathcal{Q}_\tau$, \method verifies the provisional memory state $M_{\tau}^{(0)}$ immediately after the initial construction pass.
% For efficiency, we process all probes in $\mathcal{Q}_\tau$ in parallel.
For each probe $q_j$, we retrieve top-$k$ evidence from $M_{\tau}^{(0)}$ and generate an answer with $\pi_a$:
\begin{equation}
E_j = \textsc{Search}(M_{\tau}^{(0)}, q_j), \quad
\hat{y}_j = F_{\pi_a}(q_j, E_j).
\end{equation}
A probe is considered failed if $\hat{y}_j$ is judged incorrect with respect to $y_j$.
Such failures provide localized evidence that $M_{\mathrm{0}}$ is insufficient for information introduced in or linked to $s_\tau$.

\noindent\textbf{Evidence-grounded Repair.}
For each failed probe, a reflection module converts the failure into a repair proposal.
Conditioned on the question, gold answer, predicted answer, retrieved evidence, and the provisional memory state $(q_j, y_j, \hat{y}_j, E_j, M_{\tau}^{(0)})$, it diagnoses whether the failure reflects missing information or memory content that is difficult to retrieve in its current form, and then proposes a candidate repair fact.
Collecting all failed probes in the current batch yields a set of repair proposals
\begin{equation}
    \mathcal{R}_\tau = \{r_j\}_{q_j \in \mathcal{Q}_\tau^{\mathrm{fail}}},
\end{equation}
where $\mathcal{Q}_\tau^{\mathrm{fail}} \subseteq \mathcal{Q}_\tau$ denotes the failed probes.

\noindent\textbf{Semantic Consolidation.}
Applying all repairs in $\mathcal{R}_\tau$ directly would reintroduce redundancy or conflicts, e.g., when two probes request overlapping or inconsistent additions.
We therefore consolidate the candidate repair facts against $M_{\tau}^{(0)}$.
For each candidate fact, the consolidation step assigns one of three actions with respect to the existing memory: \texttt{SKIP} if it is redundant, \texttt{MERGE} if it complements an existing entry, or \texttt{INSERT} if it is novel.
This resolves both conflicts with the existing memory and conflicts across repair proposals before any update is written back.
The refined memory is obtained as
\begin{equation}
M_{\tau}^{*} = \textsc{Refine}(M_{\tau}^{(0)}, \mathcal{R}_\tau),
\end{equation}
where $\textsc{Refine}$ denotes consolidation followed by write-back over $\mathcal{R}_\tau$.
In this way, utilization failures are detected and repaired during construction before they can propagate into later memory updates, while keeping the evolving memory compact and internally consistent.

\section{Experiments}
This section presents the experimental results.
We first compare \method with existing baselines, then evaluate 
its flexibility across storage backends, and finally assess 
the contribution of each component and key design choices.

\subsection{Experimental Setup}
\label{ssec:experimental_setup}
\noindent\textbf{Datasets.} 
We primarily evaluate \method on LoCoMo~\cite{maharana2024evaluating}, a benchmark for long-horizon conversational memory.
Following prior work~\cite{yan2025memory,fang2025lightmem}, we exclude the adversarial subset and focus on the reasoning-intensive QA setting. All LoCoMo experiments use a fixed conversation subset ($152$ QA pairs).
To further evaluate generalization beyond this controlled conversational setting, we additionally consider LongMemEval-S~\cite{jimenez2024swe}, a larger-scale long-horizon memory benchmark, and SWE-bench-Verified-mini~\cite{jimenez2024swe}, a realistic agentic coding benchmark. 
% For LongMemEval-S, we randomly sample 20\% of the examples for evaluation.
More dataset details are provided in Appendix~\ref{appendix:dataset_details}.

\noindent\textbf{Baselines.} 
We compare against two passive baselines: \emph{Full Text} and 
\emph{Naive RAG}~\cite{gao2023retrieval}, and four active memory systems: 
\emph{LangMem}~\cite{LangMemBlog}, 
\emph{Mem0}~\cite{chhikara2025mem0}, 
\emph{A-Mem}~\cite{xu2025mem}, and 
\emph{LightMem}~\cite{fang2025lightmem}.
Additional baseline details are in Appendix~\ref{appendix:baseline_details}.

\noindent\textbf{Evaluation Protocol.} Following prior work~\cite{yan2025memory,chhikara2025mem0}, we report three metrics: token-level F1 (F1), BLEU-1 (B1), and LLM-as-a-Judge accuracy (ACC).
F1 and B1 measure lexical overlap with the reference answer; 
ACC measures semantic correctness via a judge model.
GPT-4o-mini~\cite{gpt4ocard} and Claude-Haiku-4.5~\cite{anthropic2025claudehaiku45} are used as the backbones for the Memory Manager, Meta-Thinker, and Query Reasoner.
To isolate memory construction quality from answer-generation capacity, we fix GPT-4o-mini as both the Answer Agent and the LLM judge across all experiments.
The retrieval budget is top-$30$ entries, the iterative refinement budget is $H{=}3$, and we generate $J{=}5$ probe QA pairs per session for self-evolution.
Additional implementation details are in Appendix~\ref{appendix:implementation_details}.

% \noindent\textbf{Evaluation Protocol.} Following prior work~\cite{yan2025memory,chhikara2025mem0}, we report three metrics: token-level F1 (F1), BLEU-1 (B1), and LLM-as-a-Judge accuracy (ACC).
% F1 and B1 measure lexical overlap with the reference answer; 
% ACC measures semantic correctness via a judge model.
% To ensure fair comparison, we fix the retrieval budget to top-$30$ entries and keep the answer-generation setting consistent across all methods.

% \noindent\textbf{Implementation Details.}
% GPT-4o-mini~\cite{gpt4ocard} and Claude-Haiku-4.5~\cite{anthropic2025claudehaiku45} 
% are used as the default backbone for the Memory Manager, 
% Meta-Thinker, and Query Reasoner.
% The iterative query refinement budget is $H{=}3$.
% To isolate memory construction quality from answer-generation 
% capacity, we fix GPT-4o-mini as both the Answer Agent and the 
% LLM judge across all experiments.
% For in-situ self-evolution, we generate $J{=}5$ probe QA pairs 
% per session using Claude-Opus-4.5~\cite{anthropic2025claudeopus45}, 
% retrieve top-$30$ entries for verification.
% All retrieval uses text-embedding-3-small~\cite{openai2024embedding3}.

\begin{table*}[t]
\centering
\small
\caption{Results on LoCoMo across four question categories (multi-hop, temporal, open-domain, single-hop). We report F1, B1, and ACC (\%). Best results are in \textbf{bold}. GPT-4o-mini and Claude-Haiku-4.5 are backbones; GPT-4o-mini is the answer agent. $\method_{\mathrm{LM}}$ uses LightMem as storage backend.}
\label{tab:locomo_category_results}
\resizebox{\textwidth}{!}{
\begin{tabular}{c|l|ccc|ccc|ccc|ccc|ccc}
\toprule
\multirow{2}{*}{\textbf{Model}} & \multirow{2}{*}{\textbf{Method}}
& \multicolumn{3}{c|}{\textbf{Multi-Hop}}
& \multicolumn{3}{c|}{\textbf{Temporal}}
& \multicolumn{3}{c|}{\textbf{Open-Domain}}
& \multicolumn{3}{c|}{\textbf{Single-Hop}}
& \multicolumn{3}{c}{\textbf{Overall}} \\
\cmidrule(lr){3-5} \cmidrule(lr){6-8} \cmidrule(lr){9-11} \cmidrule(lr){12-14} \cmidrule(lr){15-17}
& 
& \textbf{F1} & \textbf{B1} & \textbf{ACC}
& \textbf{F1} & \textbf{B1} & \textbf{ACC}
& \textbf{F1} & \textbf{B1} & \textbf{ACC}
& \textbf{F1} & \textbf{B1} & \textbf{ACC}
& \textbf{F1} & \textbf{B1} & \textbf{ACC} \\
\midrule

\multirow{6}{*}{\textbf{\rotatebox{90}{GPT}}}
& Full Text              & 29.41 & 21.16 & 43.75 & 29.95 & 19.33 & 51.35 & 18.25 & 19.56 & 61.54 & 41.45 & 29.96 & 74.29 & 34.13 & 24.63 & 61.18 \\
& Naive RAG              & 15.84 & 9.50 & 31.25 & 17.30 & 12.36 & 35.14 & 17.40 & 16.65 & 46.15 & 39.32 & 30.35 & 58.57 & 27.14 & 20.41 & 46.05 \\
& LangMem                & 12.55 & 9.22 & 25.00 & 15.23 & 11.53 & 21.62 & 14.91 & 14.03 & 38.46 & 23.52 & 17.59 & 35.71 & 18.46 & 14.05 & 30.26 \\
& A-Mem                  & 15.56 & 10.88 & 31.25 & 55.01 & 42.40 & 51.35 & 18.18 & 15.27 & 53.85 & 42.72 & 32.43 & 62.86 & 37.90 & 28.85 & 52.63 \\
& LightMem               & 33.74 & 29.33 & 65.62 & \textbf{59.76} & \textbf{51.12} & 78.38 & \textbf{31.85} & \textbf{24.23} & \textbf{76.92} & 43.88 & 34.68 & 78.57 & 44.58 & 36.66 & 75.66 \\
& $\method_{\mathrm{LM}}$ & \textbf{48.15} & \textbf{39.67} & \textbf{78.12} & 57.21 & 41.94 & \textbf{83.78} & 24.58 & 22.44 & \textbf{76.92} & \textbf{50.45} & \textbf{38.66} & \textbf{82.86} & \textbf{49.40} & \textbf{38.28} & \textbf{81.58} \\
\midrule

\multirow{6}{*}{\textbf{\rotatebox{90}{Claude-Haiku}}}
& Full Text              & 29.41 & 21.16 & 43.75 & 29.95 & 19.33 & 51.35 & 18.25 & 19.56 & 61.54 & 41.45 & 29.96 & 74.29 & 34.13 & 24.63 & 61.18 \\
& Naive RAG              & 15.84 & 9.50 & 31.25 & 17.30 & 12.36 & 35.14 & 17.40 & 16.65 & 46.15 & 39.32 & 30.35 & 58.57 & 27.14 & 20.41 & 46.05 \\
& LangMem                & 20.05 & 14.85 & 34.38 & 34.72 & 26.33 & 37.84 & 20.01 & 20.85 & 69.23 & 22.65 & 16.19 & 48.57 & 24.81 & 18.78 & 44.74 \\
& A-Mem                  & 15.79 & 10.32 & 28.13 & 56.41 & 43.23 & 54.05 & 16.34 & 17.76 & 38.46 & 38.37 & 27.98 & 65.71 & 36.12 & 27.10 & 52.63 \\
& LightMem               & {35.11} & {31.85} & 59.38 & 58.42 & \textbf{49.85} & \textbf{89.19} & \textbf{32.60} & 24.43 & 69.23 & 44.06 & \textbf{36.56} & 71.43 & 44.69 & \textbf{37.77} & 73.03 \\
& $\method_{\mathrm{LM}}$ & \textbf{35.38} & \textbf{32.48} & \textbf{65.62} & \textbf{59.25} & 44.66 & 83.78 & 28.59 & \textbf{26.86} & \textbf{84.62} & \textbf{45.31} & 35.85 & \textbf{77.14} & \textbf{45.10} & 36.53 & \textbf{76.97} \\
\bottomrule
\end{tabular}}
% \vspace{-0.5}
\end{table*}

\subsection{Main Comparison with Baselines}
\label{ssec:evaluation_main_results}
To evaluate \method, we compare it with baselines.
We use LightMem as the storage backend of \method, denoted by $\method_{\mathrm{LM}}$. GPT-4o-mini and Claude-Haiku-4.5 are the backbones. 
Other settings follow these in Sec.~\ref{ssec:experimental_setup}.

% To evaluate the effectiveness of \method, we compare \method with baselines. We use LightMem as the storage backend of \method, denoted by $\method_{\mathrm{LM}}$.
% % LightMem is the storage backend of \method and 
% GPT-4o-mini and Claude-Haiku-4.5 are the backbones. 
% Other settings are the same as these in Sec.~\ref{ssec:experimental_setup}.

Table~\ref{tab:locomo_category_results} reports the results. Three findings emerge:
\textbf{(i)}~\emph{$\method_{\mathrm{LM}}$ achieves the best overall performance under both backbones}.
Under GPT-4o-mini, it reaches $49.40$ F1, $38.28$ B1, and $81.58$ ACC, improving over LightMem by $+4.82$ F1, $+1.62$ B1, and $+5.92$ ACC.
Under Claude-Haiku-4.5, it again achieves the best overall ACC, improving from $73.03$ to $76.97$ over LightMem.
\textbf{(ii)}~\emph{The gains are strong at the category level.}
Under GPT-4o-mini, $\method_{\mathrm{LM}}$ improves most on Multi-Hop and Single-Hop, raising ACC from $65.62$ to $78.12$ and from $78.57$ to $82.86$, respectively.
The Multi-Hop gains are consistent with diagnosis-guided iterative retrieval helping recover distributed evidence, while the Single-Hop gains suggest that construction guidance and self-evolution help preserve precise answer-bearing details.
% \textbf{(ii)}~\emph{The gains are also strong at the category level}. 
% Under GPT-4o-mini, $\method_{\mathrm{LM}}$ improves most on Multi-Hop and Single-Hop, raising ACC from $65.62$ to $78.12$ and from $78.57$ to $82.86$, respectively.
% Under Claude-Haiku-4.5, the largest improvement appears on Open-Domain, where ACC increases from $69.23$ to $84.62$.
% These patterns are consistent with the full design of $\method_{\mathrm{LM}}$, which combines reasoning-aware coordination over construction and retrieval with in-situ self-evolving memory repair. 
\textbf{(iii)}~\emph{$\method_{\mathrm{LM}}$ improves an already strong baseline.}
LightMem is already the strongest baseline, yet $\method_{\mathrm{LM}}$ further improves it under both backbones, suggesting that the gain comes from memory-cycle coordination rather than a stronger storage backend.
% We further analyze its flexibility across storage backends in Sec.~\ref{ssec:flexibility_storage}.

\subsection{Flexibility across Storage Backends}
\label{ssec:flexibility_storage}
To assess the flexibility of \method across storage backends, we instantiate it on top of three memory systems: Single-Agent~\cite{yan2025memory} ($\method_{\mathrm{SA}}$), 
A-Mem ($\method_{\mathrm{AM}}$), and 
LightMem ($\method_{\mathrm{LM}}$).
All other components and settings are fixed as in Sec.~\ref{ssec:experimental_setup}.

% To assess the flexibility of \method across storage backends, we instantiate it on top of three memory systems: a single-agent memory pipeline~\cite{yan2025memory} ($\method_{\mathrm{SA}}$), A-Mem ($\method_{\mathrm{AM}}$), and LightMem (($\method_{\mathrm{SA}}$)).
% All other components and settings are fixed as in Sec.~\ref{ssec:experimental_setup}.
% Across all settings, the Meta-Thinker, Query Reasoner, and Answer Agent are kept fixed, and only the storage backend used by the Memory Manager is changed.
% All other settings are the same as those in Sec.~\ref{ssec:experimental_setup}.

Table~\ref{tab:backend_results} reports results on LoCoMo under GPT-4o-mini.
Two observations emerge.
\textbf{(i)}~\emph{\method consistently improves all backends.}
In terms of ACC, \method improves the Single-Agent backend from $52.60$ to $84.87$, A-Mem from $52.63$ to $78.29$, and LightMem from $75.66$ to $81.58$.
For A-Mem and LightMem, the gains are also consistent in F1 and B1.
For the weaker Single-Agent backend, B1 decreases even though Acc rises sharply, suggesting that \method improves semantic correctness more than lexical overlap in this setting.
These results indicate that \method improves long-horizon memory across diverse storage implementations.
\textbf{(ii)}~\emph{The gains of \method complement storage quality rather than replace it.}
Among the enhanced variants, $\method_{\mathrm{LM}}$ achieves the strongest overall performance, which is consistent with LightMem being the strongest standalone backend.
This pattern suggests that \method improves how memory is coordinated, rather than relying on a particular storage design.

% Table~\ref{tab:backend_results} reports the results on LoCoMo with GPT-4o-mini as the backbone model. We observe: 
% \textbf{(i)}~\emph{\method consistently improves all backends.}
% For example, \method improves A-Mem from 52.63 to 78.29 ACC and 
% LightMem from 75.66 to 81.58 ACC, with gains spanning F1, B1, 
% and ACC across both backends.
% This confirms that \method's gains come from coordination over 
% the memory cycle, not from a specific storage implementation.
% \textbf{(ii)} \emph{The gains are complementary to storage quality}. Among the enhanced variants, $\method_{\mathrm{LM}}$ achieves the strongest overall performance, which is consistent with LightMem being the strongest standalone backend.
% This pattern indicates that the gains of \method complement storage quality rather than replace it.
% Stronger backends remain stronger after enhancement, while still benefiting from the proposed coordination policy.

\begin{table}[t]
\centering
\small
\caption{Flexibility across backends on LoCoMo under GPT-4o-mini. Best results per backend are in \textbf{bold}.}
\label{tab:backend_results}
{
\begin{tabular}{lccc}
\toprule
\textbf{Method} & \textbf{F1} & \textbf{B1} & \textbf{ACC} \\
\midrule
Single-Agent  & 22.64 & \textbf{17.24} & 52.60 \\
$\method_{\mathrm{SA}}$ & \textbf{23.64} & 12.94 & \textbf{84.87} \\
\midrule
A-Mem & 37.90 & 28.85 & 52.63 \\
$\method_{\mathrm{AM}}$ & \textbf{46.23} & \textbf{35.13} & \textbf{78.29} \\
\midrule
LightMem & 44.58 & 36.66 & 75.66 \\
$\method_{\mathrm{LM}}$ & \textbf{49.40} & \textbf{38.28} & \textbf{81.58} \\
\bottomrule
\end{tabular}}
\vspace{-0.5em}
\end{table}

\subsection{In-depth Dissection of MemMA}
\label{ssec:ablation}
\noindent\textbf{Ablation Studies.}
To understand the contributions of key components in \method, we implement three ablated variants on the Single-Agent backend: 
\textbf{(i)} \method/C removes Meta-Thinker guidance during construction and directly uses the Memory Manager for memory writing;
\textbf{(ii)} \method/R removes iterative retrieval, reverting to one-shot retrieval based on semantic similarity; and
\textbf{(iii)} \method/E removes the probe-and-repair loop of in-situ self-evolving memory construction and directly commits $M_{\tau}^{(0)}$ to the memory bank.
% Single-Agent is used as the storage backend. 

% GPT-4o-mini and Claude-Haiku-4.5 are used as the construction backends. 
% All other settings are the same as those in Sec.~\ref{ssec:experimental_setup}. 

\begin{figure}[t]
    \small
    \centering
    \begin{subfigure}{0.235\textwidth}
        \includegraphics[width=0.98\linewidth]{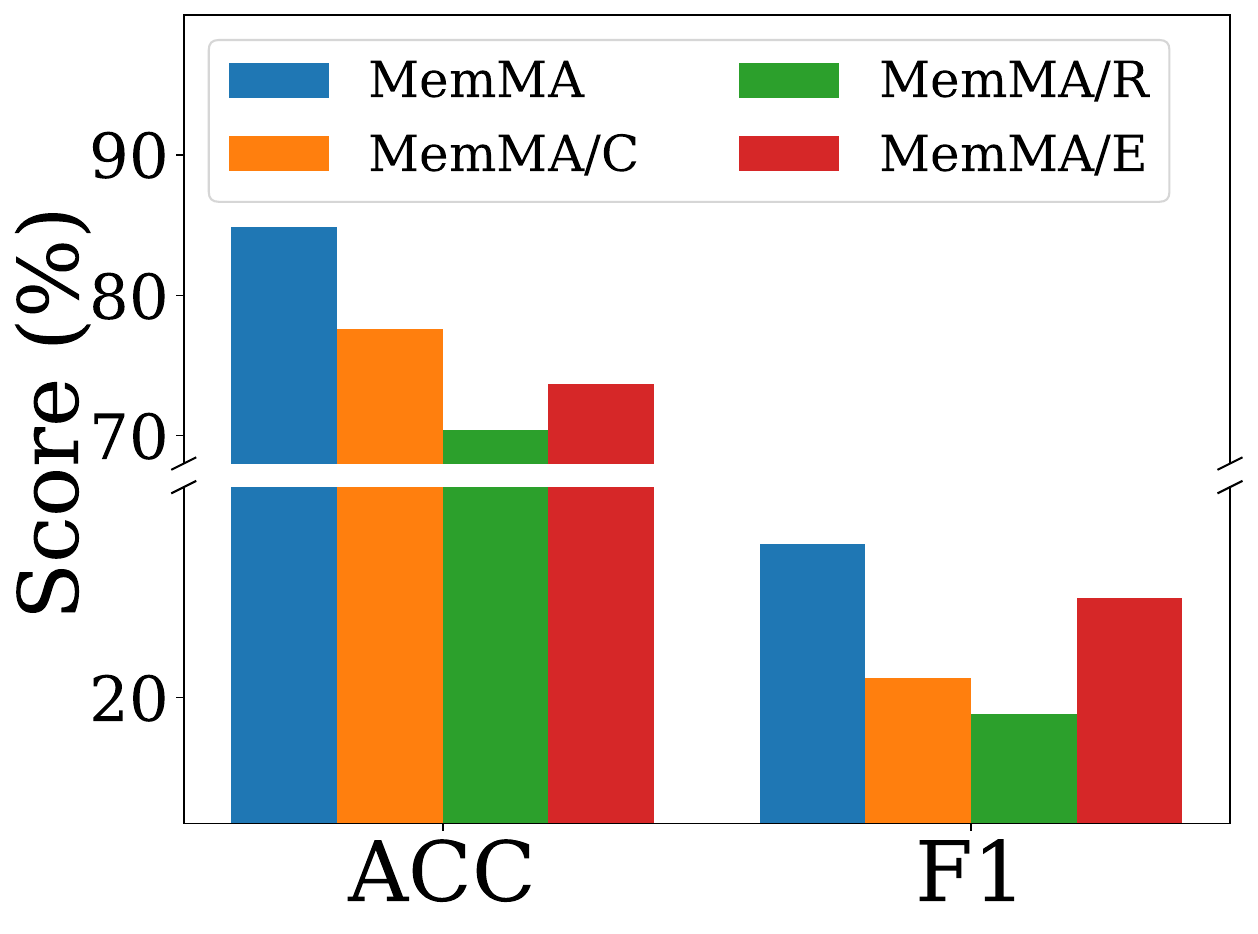}
        \vskip -0.5em
        \caption{GPT-4o-mini}
    \end{subfigure}
    \begin{subfigure}{0.235\textwidth}
        \includegraphics[width=0.98\linewidth]{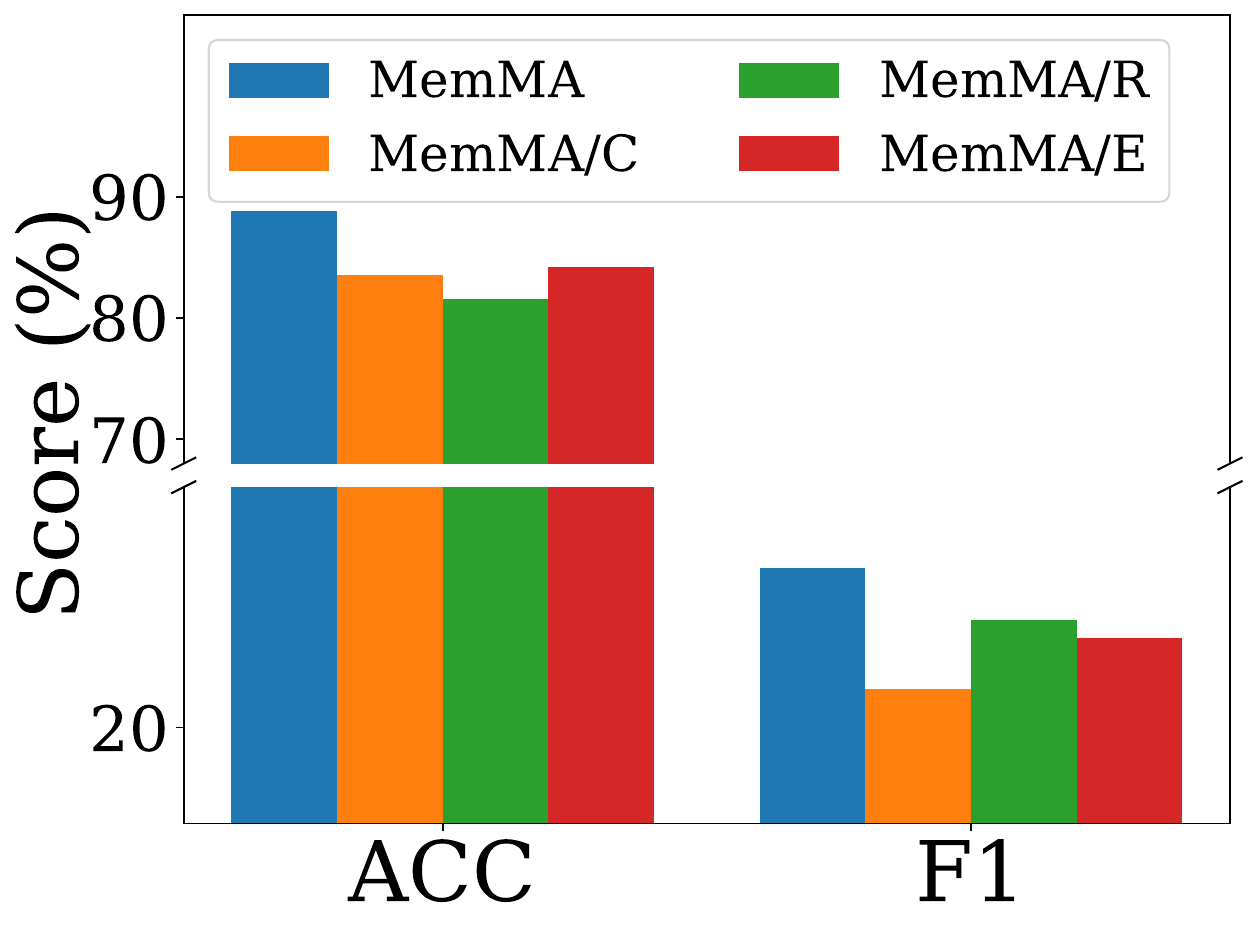}
        \vskip -0.5em
        \caption{Claude-Haiku-4.5}
    \end{subfigure}
    \vskip -0.5em
    \caption{Ablation studies of $\method_{\mathrm{SA}}$ under GPT-4o-mini and Claude-Haiku-4.5 on LoCoMo.}
    \vskip -1em
    \label{fig:ablation}
\end{figure}

Fig.~\ref{fig:ablation} reports the results under GPT-4o-mini and Claude-Haiku-4.5. The full $\method_{\mathrm{SA}}$ achieves the strongest overall performance, while the variants reveal complementary weaknesses. Specifically:
(i)~\emph{Iterative retrieval is the most critical forward-path component.}
$\method_{\mathrm{SA}}$/R causes the largest drop under both backbones, reducing ACC from $84.87$ to $70.39$ with GPT-4o-mini and from $88.82$ to $81.58$ with Claude-Haiku-4.5.
This confirms that one-shot retrieval remains a major bottleneck and that diagnosis-guided refinement is essential for narrowing the information gap.
(ii)~\emph{Self-evolution repairs construction omissions.}
$\method_{\mathrm{SA}}$/E causes the second-largest degradation 
% (ACC: $84.87$ $\to$ $73.68$ with GPT-4o-mini, $88.82$ $\to$ $84.21$ 
% with Claude-Haiku-4.5).
(ACC: $84.87$ $\to$ $73.68$ with GPT-4o-mini).
The large ACC drop with only moderate F1 change suggests that 
self-evolution mainly improves semantic correctness by repairing 
missing information during construction.
(iii)~\emph{Construction guidance reduces upstream noise.}
$\method_{\mathrm{SA}}$/C reduces ACC from  $88.82$ to $83.55$ with 
Claude-Haiku-4.5.
This shows that construction decisions benefit from explicit strategic guidance rather than local heuristics alone, as the Meta-Thinker helps determine what should be retained, consolidated, or resolved before information enters the memory bank.
These ablations confirm that \method's gains come from 
complementary improvements on both paths of the memory cycle.

\begin{figure}[t]
    \small
    \centering
    \begin{subfigure}{0.235\textwidth}
        \includegraphics[width=0.98\linewidth]{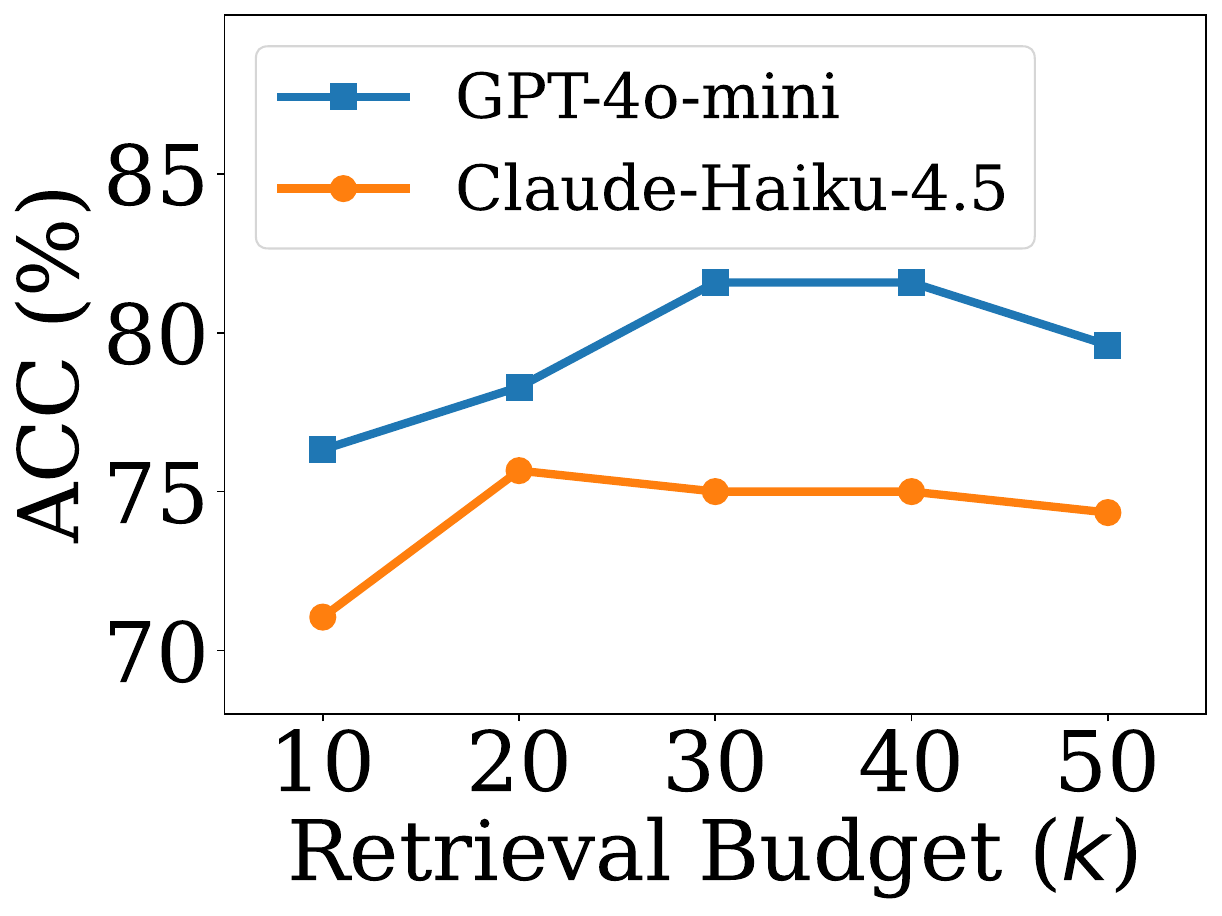}
        \vskip -0.5em
        \caption{$\method_{\mathrm{LM}}$}
    \end{subfigure}
    \begin{subfigure}{0.235\textwidth}
        \includegraphics[width=0.98\linewidth]{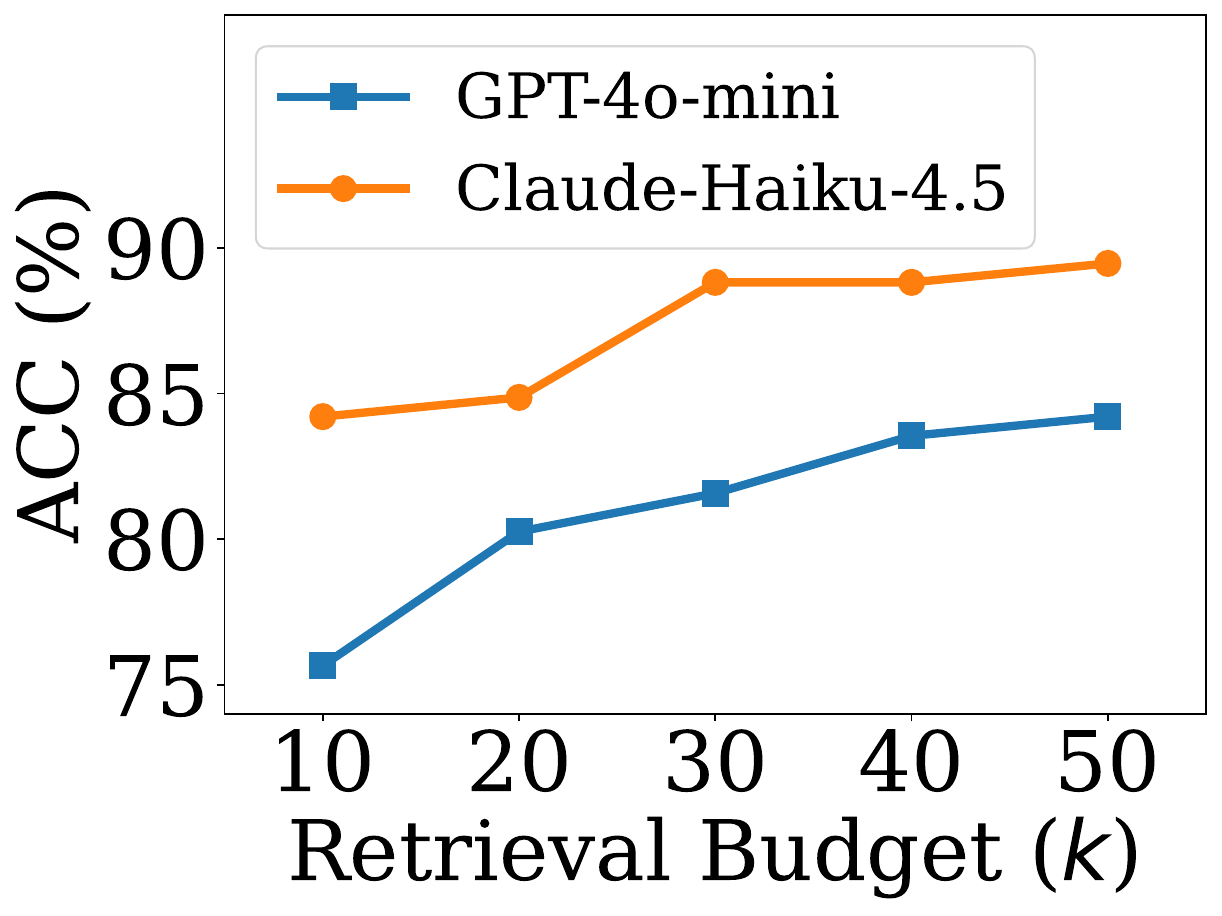}
        \vskip -0.5em
        \caption{$\method_{\mathrm{SA}}$}
    \end{subfigure}
    \vskip -0.5em
    \caption{Impact of retrieval budget $k$ of \method under both GPT-4o-mini and Claude-Haiku-4.5.}
    \vskip -1em
    \label{fig:impact_retrieval_budget}
\end{figure}

\begin{figure}[t]
    \small
    \centering
    \begin{subfigure}{0.235\textwidth}
        \includegraphics[width=0.98\linewidth]{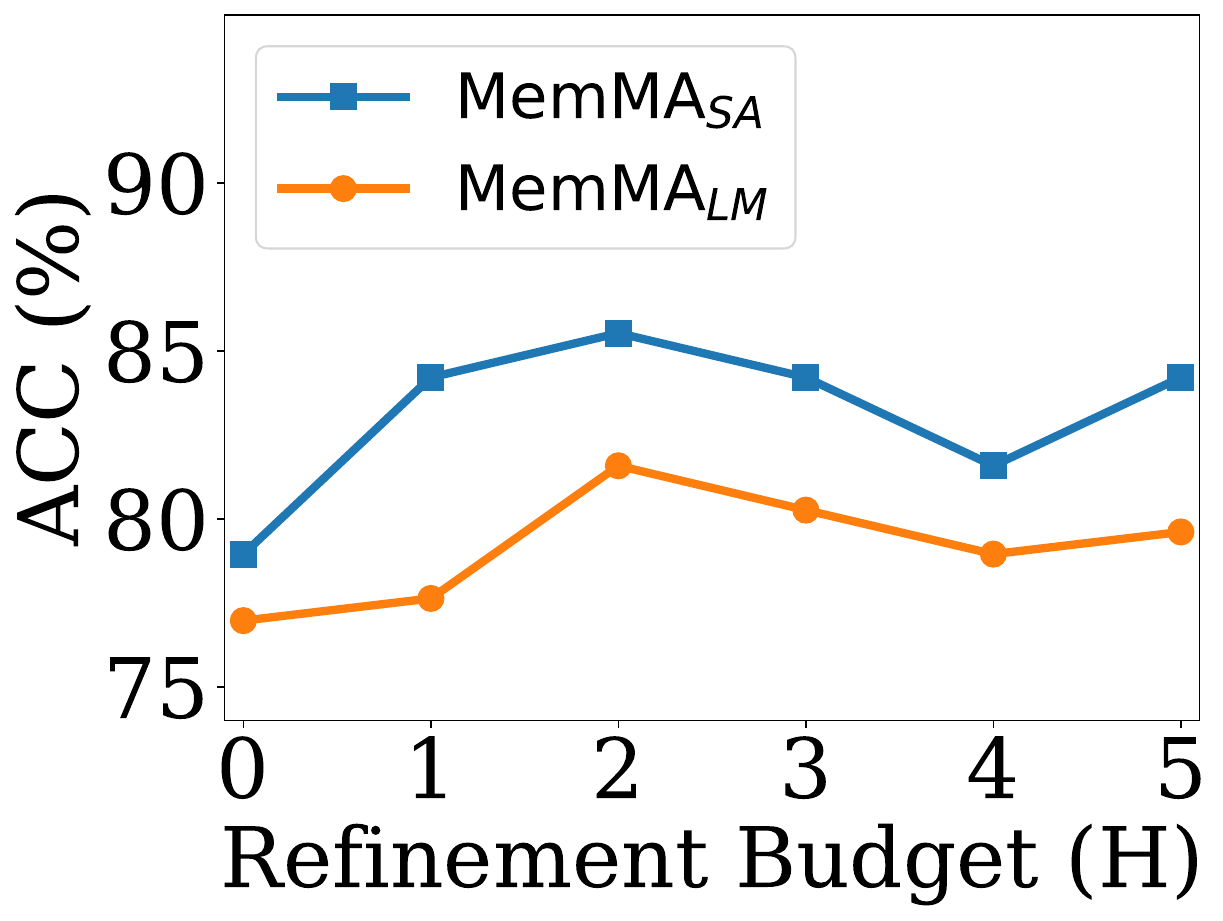}
        \vskip -0.5em
        \caption{GPT-4o-mini}
    \end{subfigure}
    \begin{subfigure}{0.235\textwidth}
        \includegraphics[width=0.98\linewidth]{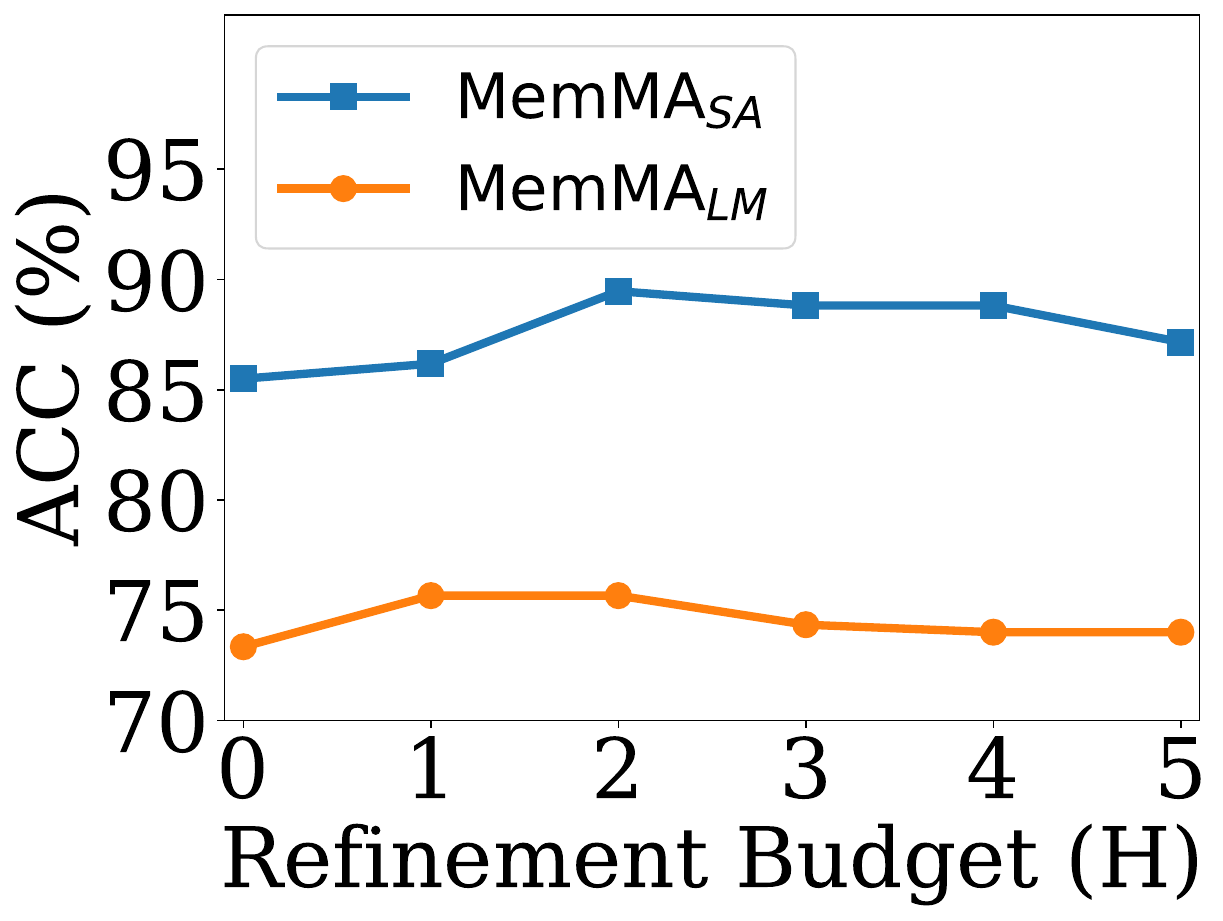}
        \vskip -0.5em
        \caption{Claude-Haiku-4.5}
    \end{subfigure}
    \vskip -0.5em
    \caption{Impact of refinement budget $H$ of \method.}
    \vskip -1em
    \label{fig:impact_retrieval_refinement}
\end{figure}

\noindent\textbf{Process-level Diagnostics.}
We directly examine whether \method mitigates the two
failure modes identified in Sec.~\ref{ssec:motivate_study}.
For \emph{Myopic Construction}, we introduce construction-process diagnostics derived from self-refinement traces. GPT-4o-mini is used as the backbone model and LightMem is used as the storage backend; all other settings follow Sec.~\ref{ssec:experimental_setup}. We measure construction gap closure, the fraction of probe-detected construction gaps repaired before memory commitment. \method closes 59/64 detected construction gaps, achieving 92.2\% gap closure. This provides direct process-level evidence that the backward path repairs construction omissions before they enter persistent memory.

For \emph{Aimless Retrieval}, we measure whether retrieval surfaces answer-bearing evidence before generation using two metrics:
\emph{evidence coverage}, the average fraction of normalized
gold-answer tokens covered by retrieved memories, and
\emph{Hit@$\geq$80\%}, the fraction of questions whose retrieved
memories cover at least 80\% of the gold-answer tokens.
For \method, we report both the initial top-$k$ evidence $E_0$ and
the final accumulated evidence $E_H$ after iterative retrieval.
Results are shown in Tab.~\ref{tab:retrieval-diagnostics}. 
\method$_{\mathrm{LM}}$ outperforms all baselines on both metrics,
while the improvement from $E_0$ to $E_H$ indicates that
diagnosis-guided iterative retrieval surfaces additional
answer-bearing evidence, providing direct process-level support for mitigating \emph{Aimless Retrieval}.

\begin{table}[t]
\centering
\small
\caption{Retrieval-process diagnostics (\%) on LoCoMo. $E_0$ and
$E_H$ denote the initial and final accumulated evidence, respectively.}
\vskip -0.5em
\setlength{\tabcolsep}{5pt}
\begin{tabular}{lcc}
\toprule
Method & Coverage & Hit@$\geq$80\% \\
\midrule
NaiveRAG & 42.2 & 24.0 \\
A-Mem & 51.0 & 26.0 \\
LightMem & 61.9 & 39.0 \\
\method$_{\mathrm{LM}}$, $E_0$ & 65.2 & 45.2 \\
\method$_{\mathrm{LM}}$, $E_H$ & \textbf{66.4} & \textbf{45.9} \\
\bottomrule
\end{tabular}
\label{tab:retrieval-diagnostics}
\end{table}

\noindent\textbf{Impact of retrieval budget $k$.} 
We vary $k \in \{10,20,30,40,50\}$ on both Single-Agent and LightMem backends and report results in Fig.~\ref{fig:impact_retrieval_budget}. We observe that the optimal $k$ depends on storage quality.
% We vary $k \in \{10,20,30,40,50\}$ with GPT-4o-mini as the backbone. Fig.~\ref{fig:impact_retrieval_budget} reports results on both Single-Agent and LightMem backends. We observe that the optimal $k$ depends on storage quality.
For $\method_{\mathrm{LM}}$, ACC peaks at $k{=}30$--$40$ ($81.58$) and declines at $k{=}50$ ($79.61$), indicating a sweet spot beyond which additional retrieval introduces noise.
For $\method_{\mathrm{SA}}$, ACC increases steadily from $75.66$ at $k{=}10$ to $84.21$ at $k{=}50$, without saturation.
We attribute this contrast to storage quality: stronger backends produce higher-quality, less redundant entries, so a moderate $k$ suffices and excess retrieval dilutes the evidence; weaker backends need a larger $k$ to retrieve enough evidence from sparser memory banks.

\noindent\textbf{Impact of retrieval refinement budget $H$.}
We vary the refinement budget $H \in \{0,1,2,3,4,5\}$ under both GPT-4o-mini and Claude-Haiku-4.5.
The results of $\method_{\mathrm{SA}}$ and $\method_{\mathrm{LM}}$ are reported in Fig.~\ref{fig:impact_retrieval_refinement}.
We observe that ACC improves sharply from one-shot retrieval ($H{=}0$) to a small $H$ and then declines.
For example, $\method_{\mathrm{SA}}$'s ACC rises from $78.95$ at $H{=}0$ to $85.53$ at $H{=}2$, then drops back to $81.58$ at $H{=}4$.
This shows that diagnosis-guided refinement converges quickly: one or two additional retrieval rounds suffice to close most information gaps, while further iterations risk retrieval drift. This validates the effectiveness of the Meta-Thinker's answerability diagnosis, which directs each refinement step toward the specific missing evidence rather than redundant searches. More analysis of the impact of probe generation model are in Appendix~\ref{appendix:impact_probe_generation_model}.

% \subsection{Case Studies}
% \label{ssec:case_studies}
% To better understand why \method improves long-horizon QA, we examine representative cases from both paths of the memory cycle.
% Our findings indicate that:
% \textbf{(i)} on the forward path, Meta-Thinker guidance helps preserve important answer-bearing details during memory construction, and diagnosis-guided iterative retrieval recovers specific facts (e.g., entity names, temporal anchors) that initial top-$k$ retrieval misses;
% \textbf{(ii)} on the backward path, 
% in-situ self-evolution converts local probe failures into targeted memory repairs that transfer to downstream QA, for example by inserting missing named entities, sharpening vague event descriptions, and completing partial evidence clusters.
% Detailed examples and traces are in Appendix~\ref{appendix:case_studies_details}.

\subsection{Case Studies}
\label{ssec:case_studies}
We conduct a case study to better understand why \method improves long-horizon QA.
% To better understand why \method improves long-horizon QA, we examine representative cases from both paths of the memory cycle.
Our findings indicate that:
\textbf{(i)} on the forward path, construction-time Meta-Thinker guidance determines whether answer-bearing details survive in memory, while diagnosis-guided iterative retrieval determines whether missing evidence is surfaced before the system commits to an answer.
Importantly, iterative retrieval cannot compensate for details that were never preserved during construction.
The cases also show that the retrieval controller and the storage backend play distinct roles: the Meta-Thinker and Query Reasoner identify the information gap, while the backend determines whether the required evidence can actually be recovered;
\textbf{(ii)} on the backward path, in-situ self-evolution converts local probe failures into targeted memory repairs that transfer to downstream QA, for example by inserting missing named entities, sharpening vague event descriptions, and completing partial evidence clusters.
Detailed examples are in Appendix~\ref{appendix:case_studies_details}.

\subsection{Generalization of MemMA}
We further evaluate whether the benefits of \method generalize beyond long-horizon conversational benchmark. We consider two complementary scenarios: a larger-scale conversational memory benchmark, LongMemEval-S~\cite{wu2025longmemeval}, and a realistic long-horizon agentic task, SWE-bench-Verified-mini~\cite{jimenez2024swe}. 

% \noindent\textbf{LongMemEval-S.} 
First, we evaluate \method on LongMemEval-S, a larger-scale long-horizon memory benchmark. We randomly sample 20\% of the examples for evaluation and compare against FullContext, NaiveRAG, A-Mem, and LightMem. Claude Haiku 4.5 is used for the Memory Manager, Meta-Thinker, and Query Reasoner, while GPT-4o-mini serves as the Answer Agent and LLM judge; other settings follow Sec.~\ref{ssec:experimental_setup}. As shown in Tab.~\ref{tab:longmemeval-s}, \method achieves 94.0\% ACC versus 82.0\% for LightMem, demonstrating that memory-cycle coordination generalizes beyond LoCoMo.

\begin{table}[t]
\centering
\small
\caption{Results on the 20\% subset of LongMemEval-S.}
\vskip -0.5em
\begin{tabular}{lc}
\toprule
Method & ACC \\
\midrule
FullContext & 50.0 \\
NaiveRAG & 58.0 \\
A-Mem & 72.0 \\
LightMem & 82.0 \\
\method$_{\mathrm{LM}}$ & \textbf{94.0} \\
\bottomrule
\end{tabular}
\label{tab:longmemeval-s}
\vskip -0.5em
\end{table}

We further evaluate \method on SWE-bench-Verified-mini to test its generalization from conversational memory to long-horizon agentic coding. Here, \method acts as a trajectory-memory controller: it organizes visible debugging trajectories into structured memories and retrieves relevant information before subsequent actions, without accessing gold patches, hidden tests, or evaluation labels. We use Claude Sonnet 4.6 for both the base coding agent and the \method controller, and compare against a vanilla agent and \method/E, which removes the probe-and-repair loop. As shown in Tab.~\ref{tab:swebench}, within-task \method improves pass rate from 62.0\% to 72.0\% (vs.\ 66.0\% for \method/E), confirming the benefit of in-situ memory repair; cross-task memory further reaches 76.0\% in the lifelong setting.

\begin{table}[t]
\centering
\caption{Results on SWE-bench-Verified-mini.}
\vskip -0.5em
\small
\begin{tabular}{llc}
\toprule
Method & Memory Scope & Pass Rate \\
\midrule
Vanilla & None & 62.0 \\
\method/E & Within-task & 66.0 \\
\method & Within-task & {72.0} \\
\method/E & Within + cross-task & 72.0 \\
\method & Within + cross-task & \textbf{76.0} \\
\bottomrule
\end{tabular}
\label{tab:swebench}
\end{table}
\section{Conclusion}
We introduce \method, a plug-and-play multi-agent framework that 
coordinates the memory cycle along its forward and backward paths.
On the forward path, a Meta-Thinker separates strategic reasoning 
from low-level execution, addressing strategic blindness in 
construction and retrieval.
On the backward path, in-situ self-evolution converts probe QA 
failures into direct memory repair before the memory is committed.
Experiments on LoCoMo show that \method outperforms all baselines 
across multiple backbones and consistently improves three different 
storage backends.

\section{Limitations}
Our evaluation focuses on a dialogue-centric long-horizon memory benchmark.
While LoCoMo covers diverse question types, including single-hop, multi-hop, temporal, and open-domain reasoning, it does not capture all settings in which persistent memory may be needed.

% Second, probe generation currently relies on a strong LLM (Claude-Opus-4.5~\cite{anthropic2025claudeopus45}); while Appendix~\ref{appendix:impact_probe_generation_model} shows that even lighter models provide meaningful repair, the use of open-source models for probe generation remains unexplored.

In addition, the backward path assumes that interaction streams can be organized into sessions and that synthetic probe QA can provide useful localized supervision.
These assumptions are natural for the benchmark studied here, but may require adaptation in settings with less clear session boundaries or more open-ended interaction structure.

\section{Ethics Statement}
This work studies long-horizon memory management for LLM agents.
All experiments are conducted on the publicly available benchmark, which consists of synthetic conversations and does not contain real user data.
No personally identifiable information is collected, stored, or processed in this work.
We note that improving memory quality in agent systems may raise broader considerations for real-world deployment, including user privacy, informed consent for data retention, controllability over stored memories, and the risk of persisting incorrect information through automated repair.
While these concerns are beyond the scope of the present study, we believe they should be treated as first-class design requirements in any production deployment of memory-augmented agents.

\section*{Acknowledgment}
This material is based upon work supported by, or in part by NSF award \#2555559. The views and conclusions contained in this material are those of the authors and should not be interpreted as necessarily representing the official policies, either expressed or implied, of the funding agencies.

\bibliography{acl_latex}

\clearpage
\appendix
\section{Full Details of Related Works}
\label{appendix:full_related_works}
\subsection{Memory-Augmented LLM Agents}
\label{appendix:memory_augmented_LLM_related_works}
External memory~\cite{hu2025memory,zhang2025survey} has become a core component of LLM agents that operate over long horizons.
Existing work can be broadly organized along three dimensions.

At the \emph{architecture level}, early systems explore how to 
structure the memory bank.
Generative Agents~\cite{park2023generative} maintains a 
chronological memory stream with reflection-based retrieval.
MemGPT~\cite{packer2023memgpt} introduces a hierarchical design 
that treats the context window as virtual memory managed by the 
LLM itself.
MemoryBank~\cite{zhong2024memorybank} adds temporal dynamics 
through forgetting-curve-based decay.
More recent work moves toward richer structure: 
SGMem~\cite{wu2025sgmem} represents dialogue as sentence-level 
graphs to capture cross-turn associations, and 
Memoria~\cite{sarin2025memoria} provides a scalable framework for 
personalized conversational memory.

At the \emph{organization level}, systems shift focus from how memory is structured to what is stored and how it is consolidated.
Mem0~\cite{chhikara2025mem0} extracts and consolidates salient facts from multi-session conversations, reducing redundancy at the source.
A-Mem~\cite{xu2025mem} goes further by dynamically organizing memories into interconnected notes following the Zettelkasten method, allowing entries to evolve as new information arrives.
LightMem~\cite{fang2025lightmem} takes a different angle, designing a lightweight multi-stage pipeline inspired by the Atkinson--Shiffrin model to balance memory quality with computational cost.
SimpleMem~\cite{liu2026simplemem} pushes efficiency further through semantic lossless compression and recursive consolidation, while EverMemOS~\cite{hu2026evermemos} introduces a self-organizing memory operating system for structured long-horizon reasoning.

At the \emph{retrieval level}, the focus shifts to how stored information is surfaced.
Zep~\cite{rasmussen2025zep} organizes memory as a temporal knowledge graph for time-aware retrieval, enabling queries that require temporal reasoning.
MemR$^3$~\cite{du2025memr} introduces a closed-loop retrieval controller with a router and an explicit evidence-gap tracker, moving retrieval from a one-shot operation to an iterative decision process.
LangMem~\cite{LangMemBlog} provides a practical SDK for memory extraction and retrieval in agent frameworks.

These methods substantially improve individual stages of the memory pipeline, but they primarily optimize storage, organization, or retrieval in isolation.
By contrast, \method addresses a broader scope: it coordinates both construction and retrieval along the forward path of the memory cycle, and further converts utilization failures into direct repair signals for the memory bank along the backward path.

\subsection{Self-Evolution and Reflection for LLM Agents}
\label{appendix:self_evolve_LLM_related_works}

A growing body of work improves LLM agents through self-feedback, while broader recent work frames persistent self-improvement as a form of agentic evolution~\cite{lin2026position,lin2026harness,liu2026adaptive,wei2026evo,wang2025survey,weng2026group,weng2024images}.
These approaches can be organized by \emph{what they modify}: the model output, an external experience store, the memory-use policy, or the memory bank itself.
Existing methods mostly operate at the first three levels; by contrast, \method directly repairs the memory bank during construction.

At the \emph{output level}, the simplest form of self-improvement operates directly on model responses.
Self-Refine~\cite{madaan2023self} iteratively critiques and revises outputs within a single generation episode, while Reflexion~\cite{shinn2023reflexion} extends this idea across episodes by storing verbal self-critiques to guide future attempts.
Similarly, TESSA~\cite{lin2024decoding} uses a reviewer agent to refine time-series annotations based on prior attempts.
These methods improve response quality, but they do not modify the underlying memory bank.

At the \emph{experience level}, systems move beyond per-episode feedback to accumulate reusable knowledge in auxiliary stores.
ExpeL~\cite{zhao2024expel} extracts natural-language insights from task trajectories and recalls them at inference time.
Voyager~\cite{wang2023voyager} builds an ever-growing skill library from environment feedback, enabling lifelong learning in open-ended settings.
O-Mem~\cite{wang2025mem} combines multiple memory types with a self-evolving mechanism for personalized agents.
These methods accumulate knowledge in separate stores, such as experience buffers or skill libraries, but do not repair entries in the primary memory bank itself.

At the \emph{policy level}, recent work improves memory management by training stronger memory-use policies through supervision, reinforcement learning, or reward optimization~\cite{guo2025deepseek,yan2025memory,lin2025comprehensive,lin2026how}.
Memory-R1~\cite{yan2025memory} trains a memory manager to learn structured operations (ADD, UPDATE, DELETE) from downstream QA supervision with sparse rewards.
Mem-$\alpha$~\cite{wang2025memalpha} extends this idea to multi-component memory systems (core, episodic, semantic), training agents to manage more complex memory architectures through interaction and feedback.
MemRL~\cite{zhang2026memrl} improves episodic memory through runtime reinforcement learning, and MEM1~\cite{zhou2025mem1} jointly optimizes memory consolidation and reasoning in an end-to-end framework.
MemBuilder~\cite{shen2026membuilder} uses synthetic QA pairs as attributed dense rewards, providing finer-grained supervision than end-task accuracy alone.
These approaches strengthen the \emph{policy} for using memory, but they still do not directly perform in-situ repair of the memory bank during construction.

In contrast, \method operates at the \emph{memory-bank level}: it directly repairs the memory bank itself during construction.
By synthesizing probe QA pairs, verifying the current memory against them, and converting failures into construction-level repair actions through evidence-grounded critique and semantic consolidation, \method provides dense, localized supervision before memory is committed, without gradient-based training or separate experience stores.

\section{Motivating Analysis Details}
\subsection{Evaluation Details}
\label{appendix:preliminary_evaluation_details}
We provide additional details for the preliminary study in Sec.~\ref{ssec:motivate_study}.

\noindent\textbf{Baseline Details.}
The three baselines are implemented by progressively enabling components of the same pipeline:
\begin{itemize}[leftmargin=*]
\item \emph{Static}: Uses a single-agent memory pipeline that processes each dialogue chunk sequentially, performs atomic memory edits (\textsc{ADD}, \textsc{UPDATE}, \textsc{DELETE}, \textsc{NONE}), and answers queries via one-shot top-$30$ retrieval based on cosine similarity. No query rewriting or strategic guidance is used.
\item \emph{Unguided Active}: Extends Static by enabling a query rewriting module that iteratively refines the retrieval query based on the retrieved evidence alone, without diagnosing what specific information is missing.
\item \emph{Strategic Active}: Further extends Unguided Active by enabling a planner that provides explicit guidance for both construction and retrieval. During construction, the planner identifies what should be retained, consolidated, or resolved. During retrieval, it diagnoses whether the current evidence is sufficient and, if not, specifies the missing information to guide the next query rewrite.
\end{itemize}

\noindent\textbf{Implementation details.}
All three baselines use GPT-4o-mini~\cite{gpt4ocard} as the backbone LLM.
The retrieval budget is top-$30$ entries.
For the two active baselines, the maximum number of query rewriting iterations is $5$.
All retrieval uses text-embedding-3-small~\cite{openai2024embedding3} for embedding. We use GPT-4o-mini as the LLM judge model for calculating ACC. The full judge prompt is shown in Table~\ref{tab:judge-prompt}.

% \noindent\textbf{LLM-as-a-Judge.}
% We use GPT-4o-mini as the judge model for ACC evaluation.
% The judge is prompted with the question, gold answer, and generated answer, and asked to label the response as \texttt{CORRECT} or \texttt{WRONG}.
% The prompt instructs the judge to be generous: as long as the generated answer touches on the same topic as the gold answer, it is counted as correct.
% For time-related questions, format differences (e.g., ``May 7th'' vs.\ ``7 May'') are accepted.
% The full judge prompt is shown in Table~\ref{tab:judge-prompt}.

\begin{table}[t]
\caption{The prompt template used for LLM-as-a-Judge evaluation.}
\begin{promptbox}[LLM-as-a-Judge Prompt]
\noindent\textbf{Task:} Label an answer to a question as \texttt{CORRECT} or \texttt{WRONG}.

\medskip
\noindent\textbf{Inputs:}
\begin{itemize}[leftmargin=*]
\item \textbf{Question:} \{question\}
\item \textbf{Gold answer:} \{gold\_answer\}
\item \textbf{Generated answer:} \{generated\_answer\}
\end{itemize}

\noindent\textbf{Instructions:}
\begin{itemize}[leftmargin=*]
\item The gold answer is usually concise. The generated answer might be longer, but be generous---as long as it touches on the same topic, count it as \texttt{CORRECT}.
\item For time-related questions, be generous with format differences (e.g., ``May 7th'' vs ``7 May'').
\item Provide a short explanation, then finish with \texttt{CORRECT} or \texttt{WRONG}.
\item Return the label in JSON: \texttt{\{``label'': ``CORRECT''\}} or \texttt{\{``label'': ``WRONG''\}}.
\end{itemize}
\end{promptbox}
\label{tab:judge-prompt}
\end{table}

\subsection{Case Studies}
\label{appendix:preliminary_case_studies}
This section provides representative examples for the preliminary study in Sec.~\ref{ssec:motivate_study}.
We organize the cases around the two pathologies of \emph{strategic blindness}.
Cases 1 and 2 illustrate \emph{Aimless Retrieval}: active rewriting alone is not sufficient if the system cannot identify what evidence is missing.
Case 3 illustrates \emph{Myopic Construction}: local memory writing may over-store low-value details or fragment one coherent episode into multiple overlapping entries.
Case 4 is a counterexample showing that \textit{Strategic Active} is not always better, because planner guidance in the current implementation is advisory rather than binding.

\noindent\textbf{Case 1: Lexical paraphrase loop in unguided retrieval.}
The question is: \emph{``When did Melanie go to the museum?''} (gold answer: \emph{5 July 2023}).
\textit{Static} misses the evidence entirely and answers \emph{``Not mentioned.''}
\textit{Unguided Active} runs five rewrite rounds, but the queries stay close to the original wording: \emph{``When did Melanie visit the museum?''}, \emph{``Melanie museum trip date''}, \emph{``Melanie's museum visit history.''}
None of these rewrites diagnose \emph{what} is missing; they only rephrase \emph{how} to ask.
The retrieved set drifts toward park, beach, and camping memories---semantically adjacent but wrong.
\textit{Strategic Active} instead identifies the gap as a missing date, notes that the evidence already contains the answer, and stops rewriting.
The first retrieved entry is the museum memory with the correct date.

\emph{Insight:} More rewrite rounds do not help if each round is a lexical paraphrase of the last.
The bottleneck is not the number of retrieval attempts but whether the system can diagnose the specific missing attribute.

\noindent\textbf{Case 2: Event ambiguity requires disambiguation, not broader search.}
The question is: \emph{``When is Caroline going to the transgender conference?''} (gold answer: \emph{July 2023}).
\textit{Unguided Active} rewrites toward increasingly generic queries: \emph{``Caroline transgender conference date''}, \emph{``Caroline upcoming events schedule''}, \emph{``Caroline future LGBTQ events.''}
The retrieved evidence mixes past LGBTQ events (e.g., a conference attended on 10 July 2023) with unrelated future activities, without resolving which conference the question refers to.
\textit{Strategic Active} narrows the gap to two specific issues: (1) the question asks about a \emph{future} conference, not a past one, and (2) \emph{transgender conference} and \emph{LGBTQ conference} may refer to different events.
One guided rewrite surfaces the relevant memory: Caroline is going to a transgender conference in July 2023.

\emph{Insight:} When the memory bank contains multiple semantically similar events, the retrieval problem is not recall but disambiguation.
Unguided rewriting broadens the search when it should narrow it.

\noindent\textbf{Case 3: Local memory writing creates filler and fragmentation.}
During construction of the early support-group conversation, \textit{Static} stores a greeting (\emph{``Caroline greeted Mel''}) as a standalone entry, then repeatedly appends details about the support-group episode to a single over-packed memory.
The result is a memory bank that mixes low-value filler with dense event summaries.
\textit{Strategic Active} partially addresses this: its planner flags information importance, temporal context, and redundancy, and even suggests consolidating similar sentiments.
However, the final memory bank still distributes the same support-group episode across several overlapping entries---attendance, emotional reaction, and self-acceptance---because the planner's guidance is only advisory and the Memory Manager still makes atomic edits one utterance at a time.

\emph{Insight:} Myopic Construction is not just about missing a planner.
Even with planning, local utterance-level editing tends to produce either filler or fragmentation, because the Memory Manager cannot perform global reorganization within a single edit step.

\noindent\textbf{Case 4: Planner guidance is advisory, not binding.}
The question is: \emph{``What activities does Melanie partake in?''} (gold answer: \emph{pottery, camping, painting, swimming}).
Here, \textit{Unguided Active} answers correctly, while \textit{Strategic Active} fails.
The planner guidance is reasonable: it suggests covering multiple activity types rather than focusing on one category.
However, the Query Reasoner judges the evidence as \textsc{answerable} and stops early.
The Answer Agent then selects a partial subset of the retrieved activities (running, reading, violin, clarinet), missing the gold-answer items entirely.

\emph{Insight:} Planner guidance in \textit{Strategic Active} is a suggestion, not a constraint.
When the downstream components ignore the guidance---by stopping retrieval too early or selecting from a biased subset of evidence---the system can still fail despite correct high-level reasoning.
This motivates the tighter coordination mechanisms in \method.

\noindent\textbf{Takeaway.}
\textit{Static} fails because one-shot retrieval often misses the evidence.
\textit{Unguided Active} adds active operators but still suffers from aimless rewriting and myopic construction.
\textit{Strategic Active} improves by diagnosing what is missing, but its guidance remains advisory: downstream components can still stop too early or select from partial evidence.
These observations motivate the design of \method, which introduces tighter coordination between the Meta-Thinker, Memory Manager, and Query Reasoner along both the forward and backward paths of the memory cycle.

\section{Meta-thinker Details}
\label{appendix:meta_thinker_details}
The Meta-Thinker $\pi_p$ produces two types of guidance: construction guidance $g_t^S$ (Sec.~\ref{ssec:architecture_design}) and retrieval guidance $g_{q,h}^R$.
The prompt for construction guidance is shown in Table~\ref{tab:prompt_construction_guidance}, and the prompt for answerability checking (which produces $g_{q,h}^R$) is shown in Table~\ref{tab:prompt_answerability}.

\section{Query Reasoner $\pi_r$ Details}
\label{appendix:query_reasoner_details}
The Query Reasoner $\pi_r$ generates the next query $u_{h+1}$ based on the Meta-Thinker's retrieval guidance $g_{q,h}^R$, as described in Sec.~\ref{ssec:architecture_design}.
The prompt is shown in Table~\ref{tab:prompt_orthogonal_query}.

\section{In-situ Self-Evolving Memory Construction Details}
\label{appendix:explicit_reflection_details}

\subsection{Synthetic QA Details}
\label{appendix:synthetic_qa_details}
After each session $s_\tau$, the system synthesizes a probe set $\mathcal{Q}_\tau = \{(q_j, y_j)\}_{j=1}^{J}$ to verify the provisional memory $M_{\tau}^{(0)}$.
% We follow the QA generation approach of MemBuilder~\cite{shen2026membuilder}, adapted to our memory structure.
% These probes are used to supervise memory construction and self-refinement, rather than serving as the final benchmark itself.
% Their role is to test whether the memory bank preserves the information required for future QA.
We group the synthetic probes into three types, each targeting a different failure mode in the memory cycle.
Table~\ref{tab:synthetic_qa_types} summarizes the taxonomy and provides one representative question--answer pair drawn from the generated probe data.

\begin{itemize}[leftmargin=*]
\item \emph{Single-hop Factoid:} Tests whether explicit facts stated in the current session $s_\tau$ are correctly stored, such as entities, attributes, or event details.
\item \emph{Multi-session Reasoning:} Tests whether the system can connect information in the current session $s_\tau$ with previously stored memory $M_{\tau-1}$, requiring cross-session integration rather than isolated fact retrieval.
\item \emph{Temporal Reasoning:} Tests whether the memory bank preserves chronological information, including relative time expressions, absolute dates, and event ordering.
\end{itemize}
\begin{table*}[h]
\centering
\small
\caption{Synthetic QA probe types used during probe generation in LoCoMo, with representative examples from the generated probe data.}
\label{tab:synthetic_qa_types}
\begin{tabular}{p{1.8cm}p{5.6cm}p{5.2cm}}
\toprule
\textbf{Type} & \textbf{Example Question} & \textbf{Example Answer} \\
\midrule
Single-hop & What type of support group did I tell Melanie I attended recently? & An LGBTQ support group \\
\midrule
Multi-hop & What is Melanie's hobby for creative expression and relaxation, and when did she create the specific piece she showed me? & Melanie paints as her hobby for creative expression and relaxation. She painted a lake sunrise last year that she showed me. \\
\midrule
Temporal & On what date and time did I have the conversation with Melanie about attending the LGBTQ support group and my career interests in counseling? & At 1:56 pm on May 8, 2023 \\
\bottomrule
\end{tabular}
\end{table*}

These synthetic probes are designed to expose common failure modes in the memory cycle, including missing entities, incomplete event details, weak cross-session linking, and temporal inconsistency.

\subsection{Prompt Details}
\label{appendix:prompt_details}
We provide the prompt templates used in the evidence-grounded repair and semantic consolidation stages of in-situ self-evolving memory construction (Sec.~\ref{ssec:self_evolving_construction}).
The probe generation stage follows the QA generation approach of MemBuilder~\cite{shen2026membuilder}, adapted to our memory structure; the probe types are described in Appendix~\ref{appendix:synthetic_qa_details}.

\noindent\textbf{Evidence-Grounded Repair.}
For each failed probe, a reflection module diagnoses whether the failure reflects missing information or content that is difficult to retrieve, and proposes a candidate repair fact $r_j$.
The prompt is shown in Tables~\ref{tab:prompt_repair} and~\ref{tab:prompt_repair_output}.

\noindent\textbf{Semantic Consolidation.}
Before writing repairs back to memory, each candidate fact is checked against existing entries and assigned one of three actions: \texttt{SKIP}, \texttt{MERGE}, or \texttt{INSERT}.
The prompt is shown in Table~\ref{tab:prompt_dedup}.

\section{Experimental Details}
\subsection{Dataset Details}
\label{appendix:dataset_details}
\noindent\textbf{LongMemEval-S.}
We evaluate on LoCoMo~\cite{maharana2024evaluating}, a benchmark for very long-term conversational memory.
LoCoMo contains $10$ conversation instances, each spanning roughly $600$ dialogue turns and $16$K tokens on average, with up to $32$ sessions.
The full benchmark includes $272$ sessions, $5{,}882$ dialogue turns, and $1{,}986$ QA pairs across the $10$ conversations.

Following prior work~\cite{yan2025memory,fang2025lightmem}, we exclude the adversarial subset and focus on the reasoning-intensive QA setting.
We use the first conversation sample (\texttt{conv-26}) as our evaluation subset.
This subset contains $19$ sessions and $419$ dialogue turns.
After excluding adversarial questions, $152$ QA pairs remain, spanning four categories: single-hop ($70$), multi-hop ($32$), temporal ($37$), and open-domain ($13$).
Using a fixed single-conversation subset ensures that all experiments and ablations are performed on exactly the same conversation and evaluation set.

\noindent\textbf{LongMemEval-S.}
LongMemEval~\cite{wu2025longmemeval} evaluates the long-term interactive memory of chat assistants with $500$ questions posed over multi-session chat histories, covering information extraction, multi-session reasoning, temporal reasoning, knowledge updates, and abstention.
The S variant provides, for each question, a chat history of roughly $115$K tokens, which is substantially longer than a single LoCoMo conversation.
We randomly sample $20\%$ of the questions ($100$ questions) as our evaluation set.

\noindent\textbf{SWE-bench-Verified-mini.}
SWE-bench~\cite{jimenez2024swe} asks an agent to resolve real GitHub issues by producing a patch that passes hidden tests. SWE-bench Verified is a $500$-task human-validated subset. We use SWE-bench-Verified-mini, a $50$-task subset selected to preserve the performance, test-pass-rate, and difficulty distribution of the full Verified set while requiring far less storage.

\subsection{Baseline Details}
\label{appendix:baseline_details}
We compare \method against both passive and active baselines:
\begin{itemize}[leftmargin=*]
    \item \textbf{Full Text}: concatenates the entire dialogue history into the context window and answers directly without memory construction or retrieval.
    \item \textbf{Naive RAG}~\cite{gao2023retrieval}: splits the dialogue into fixed-size chunks, embeds them, and retrieves the top-$k$ chunks by cosine similarity at query time.
    \item \textbf{LangMem}~\cite{LangMemBlog} provides a practical SDK for memory extraction and retrieval in agent frameworks, storing memories as structured key-value entries.
% \item \textbf{Mem0}~\cite{chhikara2025mem0} extracts and consolidates salient facts from multi-session conversations, reducing redundancy at the source through deduplication and merging.
\item \textbf{A-Mem}~\cite{xu2025mem} dynamically organizes memories into interconnected notes following the Zettelkasten method, allowing entries to evolve as new information arrives through activation-based retrieval.
\item \textbf{LightMem}~\cite{fang2025lightmem} designs a lightweight multi-stage pipeline inspired by the Atkinson--Shiffrin model, organizing memory into sensory, short-term, and long-term stores to balance quality with computational cost.

\end{itemize}

\subsection{Implementation Details}
\label{appendix:implementation_details}
% \noindent\textbf{Implementation Details.}
GPT-4o-mini~\cite{gpt4ocard} and Claude-Haiku-4.5~\cite{anthropic2025claudehaiku45} 
are used as the default backbone for the Memory Manager, 
Meta-Thinker, and Query Reasoner.
The iterative query refinement budget is $H{=}3$.
To isolate memory construction quality from answer-generation 
capacity, we fix GPT-4o-mini as both the Answer Agent and the 
LLM judge across all experiments.
For in-situ self-evolution, we generate $J{=}5$ probe QA pairs 
per session using Claude-Opus-4.5~\cite{anthropic2025claudeopus45}, 
retrieve top-$30$ entries for verification.
All retrieval uses text-embedding-3-small~\cite{openai2024embedding3}.
%%%%%%%%%%%%%%%%%%%%%%%%%%%%%%%%%%%%%%%%%%%%%%%%%%%%%%%%%%%%%%%%%%%%%%%%%%%%%%

% \section{Addition Results of Flexibility across Storage Backends}
% \label{appendix:more_results_flexibility}

% \section{Addition Results of In-depth Dissection of MemMA}
% \label{appendix:in_depth_dissection}
\section{Impact of Probe Generation Model}
\label{appendix:impact_probe_generation_model}

\begin{table}[t]
\centering
\small
\caption{Impact of probe generation model on $\method_{\mathrm{LM}}$ with Claude-Haiku-4.5 as the construction backbone. Best results are in \textbf{bold}.}
\label{tab:probe_generation}
\begin{tabular}{lccc}
\toprule
\textbf{Probe Model} & \textbf{F1} & \textbf{B1} & \textbf{ACC} \\
\midrule
Claude-Haiku-4.5   & 44.98 & \textbf{35.69} & 74.34 \\
Claude-Sonnet-4.5  & 43.30 & 32.74 & 74.34 \\
Claude-Opus-4.5    & \textbf{45.10} & 35.66 & \textbf{76.97} \\
\bottomrule
\end{tabular}
\end{table}

\subsection{Empirical Analysis.}
To understand how probe quality affects in-situ self-evolving memory construction (Sec.~\ref{ssec:self_evolving_construction}), we vary the probe generation model among Claude-Haiku-4.5, Claude-Sonnet-4.5~\cite{anthropic2025claudesonnet45}, and Claude-Opus-4.5.
$\method_{\mathrm{LM}}$ with Claude-Haiku-4.5 as the construction backbone is used. All other settings follow Sec.~\ref{ssec:experimental_setup}.

Table~\ref{tab:probe_generation} reports the results. We observe that:
\textbf{(i)}~\emph{Opus achieves the best overall repair quality.}
It reaches $76.97$ ACC and $45.10$ F1, outperforming both Haiku ($74.34$ ACC, $44.98$ F1) and Sonnet ($74.34$ ACC, $43.30$ F1).
\textbf{(ii)}~\emph{Haiku and Sonnet match in ACC but diverge in lexical metrics.}
Despite identical ACC, Haiku outperforms Sonnet in F1 ($44.98$ vs.\ $43.30$) and B1 ($35.69$ vs.\ $32.74$), indicating that Haiku's probes lead to higher-quality memory repairs at the token level.

We attribute this gap to differences in probe style.
Sonnet tends to produce shorter, more extractive QA pairs (average answer length $11.12$ words, with $136$ out of $380$ answers containing $\leq 3$ words), while Haiku generates longer probes (average answer length $19.43$ words) with more multi-session and temporal-reasoning questions.
Opus produces probes of moderate length (average answer length $21.48$ words) with the highest proportion of cross-session relational questions.
Overly short probes test only surface-level keyword recall rather than cross-session consistency, so they provide weaker signals for diagnosing and repairing construction omissions.

\subsection{Qualitative Examples.}
To better understand the performance gap, we analyze the probe statistics and show representative examples in Table~\ref{tab:probe_statistics} and Table~\ref{tab:probe_examples}.

\begin{table*}[h]
\centering
\small
\caption{Probe statistics across generation models. ``One-word / $\leq$3'' counts one-word and short ($\leq 3$ words) answers out of $95$ total per model. Question type counts follow the taxonomy in Appendix~\ref{appendix:synthetic_qa_details}.}
\label{tab:probe_statistics}
\begin{tabular}{lcccccc}
\toprule
\textbf{Probe Model} & \textbf{Avg. Q Len.} & \textbf{Avg. A Len.} & \textbf{One-word / $\leq$3} & \textbf{Single-hop} & \textbf{Multi-hop} & \textbf{Temporal} \\
\midrule
Haiku   & 18.48 & 19.44 & 4 / 15  & 55 & 25 & 15 \\
Sonnet  & 15.42 & 11.13 & 11 / 33 & 64 & 16 & 15 \\
Opus    & 17.38 & 21.55 & 4 / 8   & 58 & 26 & 11 \\
\bottomrule
\end{tabular}
\end{table*}

\begin{table*}[t]
\centering
\small
\caption{Representative probe QA pairs from the same dialogue session. Sonnet's single-word answer tests only keyword presence, while Haiku and Opus require multi-attribute recall.}
\label{tab:probe_examples}
\begin{tabular}{p{1.2cm}p{5.5cm}p{5.5cm}}
\toprule
\textbf{Model} & \textbf{Question} & \textbf{Answer} \\
\midrule
Haiku & What has the support group I attended done for my personal development and self-acceptance? & The support group has made me feel accepted and given me courage to embrace myself. \\
\midrule
Sonnet & What did the LGBTQ support group help me feel that gave me courage to embrace myself? & Accepted \\
\midrule
Opus & How has attending the LGBTQ support group influenced my personal growth and willingness to be open about my identity? & The support group has been a safe space that made me feel accepted, giving me the courage to embrace myself and be more open about my identity in other areas of life. \\
\bottomrule
\end{tabular}
\end{table*}

Two patterns stand out.
First, Sonnet generates significantly more short answers: $11$ one-word and $33$ answers with $\leq 3$ words, compared to $4$ / $15$ for Haiku and $4$ / $8$ for Opus.
Sonnet's probes tend to compress answers into factoid-style keywords (e.g., ``Accepted''), which tests keyword presence but not whether the memory bank can support multi-attribute reasoning.
The issue is not that Sonnet hallucinates, but that it loses information by over-compressing, resulting in weaker supervision for memory repair.

Second, Sonnet's probes are dominated by single-hop questions ($64$ out of $95$), while Haiku and Opus allocate more probes to multi-hop reasoning ($25$ and $26$, respectively).
Since single-hop probes only verify whether individual facts were stored, they are less likely to expose consolidation failures where information from different sessions was not properly linked.
The higher proportion of multi-hop probes in Haiku and Opus explains their stronger repair quality.

Sonnet's single-word answer (``Accepted'') only checks whether the memory bank contains a specific keyword.
Haiku and Opus instead require the memory to support reasoning over multiple attributes (personal development, self-acceptance, courage), which is more likely to reveal gaps in cross-session consolidation.
This explains why Sonnet, despite matching Haiku in ACC, falls behind in lexical metrics: its probes trigger fewer and shallower repairs.

\section{Full Details of Case Studies of MemMA}
\label{appendix:case_studies_details}
In this section, we expand the details of case studies in Sec.~\ref{ssec:case_studies}.
We organize the cases by the two paths of the memory cycle.
For the forward path, we separately examine construction-time Meta-Thinker guidance (Sec.~\ref{ssec:architecture_design}) and iterative query refinement.
For the backward path, we examine how in-situ self-evolving memory construction (Sec.~\ref{ssec:self_evolving_construction}) repairs the memory bank with facts that later improve downstream benchmark QA.

\subsection{Forward Path: Construction-Time Meta-Thinker Guidance}
\label{appendix:case_construction_guidance}

To isolate the effect of construction-time meta guidance, we compare $\method_{\mathrm{SA}}$ against the ablated variant $\method_{\mathrm{SA}}$/C using Claude-Haiku-4.5 as the construction backbone.
Both variants share the same query-time components, including answerability diagnosis and iterative query refinement; the only difference is whether the Meta-Thinker provides construction guidance $g_t^S$ to the Memory Manager.

\noindent\textbf{Case 1: Preserving answer-bearing visual detail.}
Consider the question: \emph{``What did Caroline find in her neighborhood during her walk?''}
$\method_{\mathrm{SA}}$ answers \emph{``Caroline came across a rainbow sidewalk ...''}, whereas $\method_{\mathrm{SA}}$/C produces a vague answer about \emph{``cool stuff''} and even confuses the walking event with a biking outing.

According to the construction trajectory,  
with guidance enabled, the Meta-Thinker's construction guidance $g_t^S$ explicitly lists the answer-bearing visual object \emph{rainbow sidewalk}, together with its supporting attributes such as \emph{Pride Month} and \emph{cool / vibrant / welcoming}.
The Memory Manager then stores a clean entry containing the exact answer-bearing detail.
Without guidance, this object detail is absent from the memory bank, so later retrieval can only recover semantically adjacent but insufficient context.
This case shows that construction-time guidance preserves concrete object-level details that iterative query refinement cannot recover once they are lost.

\noindent\textbf{Case 2: Preventing destructive merges.}
The question \emph{``What instruments does Melanie play?''} reveals a different failure mode.
$\method_{\mathrm{SA}}$ correctly answers \emph{``the clarinet and the violin,''} whereas $\method_{\mathrm{SA}}$/C answers \emph{``the clarinet''} and even incorrectly claims that Melanie does not play the violin.

The key difference lies in the constructed memory.
With guidance, the Memory Manager stores the clarinet and violin facts as distinct entries, preserving them as parallel details.
Without guidance, the Memory Manager incorrectly merges them into a conflicting entry, effectively overwriting one fact with another.
This case shows that construction-time guidance also prevents harmful consolidation that would later produce factually incorrect retrieval results.

\noindent\textbf{Takeaway.}
These cases show that the Meta-Thinker's construction guidance $g_t^S$ improves the memory bank before retrieval begins.
In particular, it preserves exact answer-bearing details, keeps semantically adjacent facts disentangled, and avoids destructive merges that would otherwise create retrieval drift or contradictions.
Additional examples, including quoted textual details (\emph{``trans lives matter''}) and topic disentanglement (\emph{adoption research} vs.\ \emph{counseling research}), follow the same pattern.

\subsection{Forward Path: Iterative Query Refinement}
\label{appendix:case_forward_refinement}
The second part of the forward path is Meta-Thinker-guided iterative retrieval.
Here, retrieval operates over a fixed memory bank; the Meta-Thinker first judges whether the current evidence is sufficient (\textsc{answerable} vs.\ \textsc{not-answerable}), and the Query Reasoner then refines the query to retrieve the missing evidence.

\paragraph{Case 1: Recovering a temporal anchor.}
Consider the question: \emph{``When did Caroline go to the LGBTQ conference?''}
The Single-Agent baseline answers \emph{``Not mentioned in the conversation,''} treating the information gap as an absence of information.
By contrast, $\method_{\mathrm{SA}}$ first judges the current evidence as \textsc{not-answerable}, noting that the problem is not the absence of all related memories, but the lack of an exact date and the ambiguity between \emph{LGBTQ conference} and \emph{transgender conference}.
The Query Reasoner then issues increasingly targeted queries, such as asking for the specific date in July 2023 and explicitly disambiguating the two event names.
The final answer becomes \emph{``July 10, 2023.''}

This case shows that the forward path does not improve performance by making better guesses; it improves performance by delaying commitment until the temporal anchor is retrieved.

\noindent\textbf{Case 2: Filling a missing entity.}
A second example concerns the question: \emph{``Where did Caroline move from 4 years ago?''}
The LightMem baseline answers \emph{``Her home country,''} which is directionally correct but incomplete because the benchmark expects the country name.
$\method_{\mathrm{LM}}$ judges the evidence as \textsc{not-answerable}: the relation is already known but the specific entity is missing.
The Query Reasoner then rewrites the query around this information gap, first asking about Caroline's home country before she moved four years ago and then asking more explicitly for the country name.
The final answer becomes \emph{``Her home country, Sweden.''}

This case is informative because the same diagnostic pattern also appears with the weaker Single-Agent backend.
There, the Meta-Thinker correctly identifies the same information gap, but the backend does not contain the relevant entry.
Thus, the Meta-Thinker and Query Reasoner can accurately locate the gap regardless of backend, but the final answer depends on whether the memory bank contains the answer-bearing entry.

\paragraph{Case 3: Recovering a missing event detail.}
For the question \emph{``What did Melanie and her family see during their camping trip last year?''}, the baseline answers \emph{``They saw amazing views,''} which is too generic to be judged correct.
$\method_{\mathrm{LM}}$ instead judges the evidence as \textsc{not-answerable}, performs one additional refinement round, and recovers the specific answer \emph{``Perseid meteor shower.''}
The key point here is that the answer already exists in the memory bank; the initial top-$k$ retrieval simply failed to surface the decisive detail.
Iterative refinement fixes this by turning a vague event description into a concrete answer.

\paragraph{Takeaway.}
Across these cases, the Meta-Thinker first identifies the information gap---a temporal anchor, a missing entity, or a specific event detail---and the Query Reasoner translates that gap into a more targeted retrieval query.
The forward-path gain therefore comes not from stronger answer generation, but from refusing to answer too early and iteratively retrieving until the information gap is closed.

\subsection{Backward Path: In-Situ Self-Evolving Memory Construction}
\label{appendix:case_backward}
To isolate the effect of in-situ self-evolution, we compare the full $\method_{\mathrm{SA}}$ against the ablated variant $\method_{\mathrm{SA}}$/E using GPT-4o-mini as the construction backbone.
Both variants share the same construction-time Meta-Thinker guidance and query-time components; the only difference is whether the probe-and-repair loop (Sec.~\ref{ssec:self_evolving_construction}) is applied after each session.
The following cases show that self-evolution improves performance not only by improving probe QA accuracy, but by writing back repair facts that later change downstream benchmark answers from incorrect to correct.

\noindent\textbf{Case 1: Named-entity insertion for concert-related QA.}
During self-evolution of session $\tau{=}10$, the probe \emph{``What is the name of the artist who performed at Melanie's daughter's birthday concert?''} fails.
Before self-evolution, the system answers that the artist is not mentioned in memory; after self-evolution, it answers \emph{``Matt Patterson.''}
The repair trace shows that self-evolution inserts the following candidate repair fact:
\begin{quote}
\small
\texttt{ADD\_FACT}: ``The artist who performed at Melanie's daughter's birthday concert is Matt Patterson.''
\end{quote}
A related repair later adds another musical entity, \emph{Summer Sounds}.

These inserted facts directly transfer to the downstream benchmark question \emph{``What musical artists/bands has Melanie seen?''}
Without self-evolution, the system answers only that \emph{``a band performed at a show''} but cannot name it.
With self-evolution, the answer becomes \emph{``Summer Sounds''} and \emph{``Matt Patterson.''}
Probe failures expose that the memory bank contains event descriptions but not the exact entity names required for downstream QA.

\noindent\textbf{Case 2: Restoring a distinctive event detail.}
During self-evolution, the probe \emph{``What was Melanie's most memorable camping experience with her family?''} fails.
The system produces a generic answer about roasting marshmallows and telling stories, missing the distinctive detail.
Self-evolution repairs this by inserting a new event fact centered on the Perseid meteor shower.

This repair transfers to the downstream benchmark question \emph{``What did Melanie and her family see during their camping trip last year?''}
Without self-evolution, the downstream answer remains generic and mentions only ordinary camping activities.
With self-evolution, the system retrieves and outputs the specific event detail \emph{``Perseid meteor shower.''}
This case shows that self-evolution sharpens vague event memories into distinctive, answerable ones.

\noindent\textbf{Case 3: Completing a partial evidence cluster.}
During self-evolution, the probe \emph{``What new pottery project did Melanie recently finish, and what was her earlier pottery creation?''} fails.
The system can only answer part of the question and leaves the pottery objects underspecified.
Self-evolution repairs this by writing back the missing facts about a \emph{colorful bowl} and an earlier \emph{black and white bowl}.

These repairs transfer to downstream benchmark questions such as \emph{``What types of pottery have Melanie and her kids made?''} and \emph{``What kind of pot did Mel and her kids make with clay?''}
Without self-evolution, the model answers only with generic descriptions such as \emph{``pots''} or \emph{``various pottery projects.''}
With self-evolution, the final answer becomes object-level and complete: bowls, a cup with a dog face, a colorful bowl, and a black-and-white bowl.
This case illustrates that self-evolution does not only insert isolated facts; it can also complete a sparse local evidence cluster so that the whole topic becomes answerable.

\noindent\textbf{Takeaway.}
Across these cases, in-situ self-evolution improves performance by turning vague, generic, or partially correct memory regions into retrieval-friendly, answerable memory units.
More specifically, it works through three recurring repair mechanisms:
(i) named-entity insertion,
(ii) distinctive event-detail sharpening, and
(iii) partial evidence completion.
The key point is that probe failures do not remain local.
Instead, they are converted into evidence-grounded repair actions that transfer directly to downstream benchmark performance.

\begin{table}[t]
\caption{The prompt template used for Meta-Thinker construction guidance.}
\begin{promptbox}[Meta-Thinker Construction Guidance Prompt]
\noindent\textbf{Role:} You are a quality-control checker for a memory construction system. Given one conversation utterance, list every distinct factual statement it contains. Each fact must be an atomic, self-contained statement that could answer a WHO/WHAT/WHEN/WHERE/HOW MANY question.

\medskip
\noindent\textbf{Rules:}
\begin{itemize}[leftmargin=*]
\item Extract EVERY fact---do not skip anything. Err on the side of over-extraction.
\item Use the speaker's exact words for names, objects, dates, places, and quantities.
\item One fact per line. Do NOT merge multiple facts into one line.
\item Prefix each fact with the correct speaker name.
\item Do NOT interpret emotions, themes, values, or symbolism.
\item Do NOT paraphrase---preserve the original phrasing.
\end{itemize}

\medskip
\noindent\textbf{Output format:}\\
\texttt{FACTS:}\\
\texttt{- [Speaker] fact 1}\\
\texttt{- [Speaker] fact 2}\\
\texttt{- ...}
\end{promptbox}
\label{tab:prompt_construction_guidance}
\end{table}

\begin{table}[t]
\caption{The prompt template used for Meta-Thinker answerability checking (Part 1).}
\begin{promptbox}[Meta-Thinker Answerability Checking Prompt (Part 1)]
\noindent\textbf{Role:} You are a Meta-Thinker agent for answerability checking in a memory-augmented QA system.

\medskip
\noindent\textbf{Inputs:}
\begin{itemize}[leftmargin=*]
\item Question
\item Retrieved memories grouped by speaker (memory\_id, timestamp, snippet)
\item Previous queries
\end{itemize}

\noindent\textbf{Goal:} Minimize false \texttt{NOT\_ANSWERABLE} while staying evidence-grounded.

\medskip
\noindent\textbf{Blocking-gap test:} Return \texttt{NOT\_ANSWERABLE} only if a missing fact or unresolved contradiction would CHANGE the final short answer. If a best-supported answer is already stable, return \texttt{ANSWERABLE}.

\medskip

\noindent\textbf{Granularity policy:}
\begin{itemize}[leftmargin=*]
\item Time questions: require exact day/date only if the question explicitly asks for it; otherwise accept the best unambiguous granularity.
\item Who/what/which: one clearly supported entity is enough unless the question explicitly requests exhaustive output.
\item Contradictions: only contradictions that change the final answer are blocking.
\end{itemize}

% \medskip
\end{promptbox}
\label{tab:prompt_answerability}
\end{table}

\begin{table}[t]
\caption{The prompt template used for Meta-Thinker answerability checking (Part 2).}
\begin{promptbox}[Meta-Thinker Answerability Checking Prompt (Part 2)]
% \medskip
\noindent\textbf{Anti-stall:} If $\geq$3 previous queries were attempted and the same non-blocking gap repeats, prefer \texttt{ANSWERABLE} at best-supported granularity.

\medskip

\noindent\textbf{Output format:}

\texttt{<decision>ANSWERABLE|NOT ANSWERABLE</decision>}

\texttt{<reason>1--3 sentences about the asked slot only.</reason>}
\texttt{<key-gaps>Ranked bullets if NOT-ANSWERABLE; NONE otherwise.</key-gaps>}\\
\texttt{<missing-speaker> speaker-1 | speaker-2 | both | unknown</missing-speaker>}

\texttt{<time-need>Required granularity and missing anchor, or NONE.</time-need>}\\
\texttt{<retrieval-guidance>Goal, suggested queries, keywords, constraints, avoid terms.</retrieval-guidance>}
\end{promptbox}
\label{tab:prompt_answerability_output}
\end{table}

\begin{table}[t]
\caption{The prompt template used for Query Reasoner $\pi_r$ to generate orthogonal query $u_{h+1}$.}
\begin{promptbox}[Orthogonal Query Generation Prompt]
\noindent\textbf{Role:} You are an expert Query Rewriter for conversation memory retrieval.

\medskip
\noindent\textbf{Inputs:}
\begin{itemize}[leftmargin=*]
\item The question and Meta-Thinker diagnosis (top gap, missing speaker, time need, constraints, avoid terms)
\item Previous queries and retrieval trace (query $\to$ retrieved memory IDs)
\end{itemize}

\noindent\textbf{Task:} Generate EXACTLY ONE new retrieval query that targets the top gap and is maximally likely to retrieve new evidence.

\medskip
\noindent\textbf{Hard rules:}
\begin{itemize}[leftmargin=*]
\item Do NOT repeat any previous query verbatim or near-verbatim.
\item MUST target the top gap only (do not broaden to multiple gaps).
\item MUST include all constraints (entity + time/version) exactly as provided.
\item If time need is provided, include both the relative phrase (e.g., ``last year'') and the computed absolute time (e.g., ``2021'').
\item MUST avoid exhausted terms.
\item Prefer disambiguation queries if contradiction exists.
\item If missing speaker is specified, phrase the query to target that speaker's perspective.
\end{itemize}

\medskip
\noindent\textbf{Output format (JSON):}\\
\texttt{\{``rewritten\_query'': ``...'', ``strategy'': ``...'', ``target\_speaker'': ``...''\}}
\end{promptbox}
\label{tab:prompt_orthogonal_query}
\end{table}

\begin{table}[h]
\caption{The prompt template used for evidence-grounded repair in self-evolution (Part 1).}
\begin{promptbox}[Evidence-Grounded Repair Prompt (Part 1)]
\noindent\textbf{Role:} You are a memory-repair assistant for a two-speaker conversation memory system.

\medskip
\noindent\textbf{Inputs:}
\begin{itemize}[leftmargin=*]
\item Question, Gold Answer, Model Answer
\item Retrieved evidence snippets (from current memory; may be irrelevant if the info is missing)
\end{itemize}

\noindent\textbf{Task:} Decide whether to add one fact to memory so the system can answer correctly next time.

\medskip
\noindent\textbf{Decision rules (priority order):}
\begin{itemize}[leftmargin=*]
\item If Gold Answer is unanswerable, output \texttt{NOOP}.
\item If Gold Answer is answerable and Model Answer is wrong or incomplete, output \texttt{ADD\_FACT}. The fact should capture the key information from the Gold Answer.
\item If Gold Answer and Model Answer are essentially equivalent, output \texttt{NOOP}.
\end{itemize}

\medskip
\noindent\textbf{Quality rules:}
\begin{itemize}[leftmargin=*]
\item Fact must be concrete, specific, and retrieval-friendly (include names, dates, details).
\item Preserve relative date expressions verbatim; do NOT convert to absolute dates.
\item Assign \texttt{target\_speaker} based on who the fact is about.
\end{itemize}
\end{promptbox}
\label{tab:prompt_repair}
\end{table}

\begin{table}[h]
\caption{The prompt template used for evidence-grounded repair in self-evolution (Part 2: output format and examples).}
\begin{promptbox}[Evidence-Grounded Repair Prompt (Part 2)]
\noindent\textbf{Output format (JSON):}\\
\texttt{\{``op'': ``ADD\_FACT | NOOP'', ``target\_speaker'': ``speaker\_a | speaker\_b'', ``fact'': ``...'', ``evidence\_span'': ``...'', ``confidence'': 0.0, ``reason'': ``...''\}}

\medskip
\noindent\textbf{Example --- information missing from memory:}\\
\texttt{Question:} What does the necklace from Caroline's grandma symbolize?\\
\texttt{Gold:} Love, faith, and strength.\\
\texttt{Model:} The memories do not contain information about a necklace.\\
\texttt{Output:} \texttt{\{``op'': ``ADD\_FACT'', ``fact'': ``Caroline's grandma gave her a necklace from Sweden that symbolizes love, faith, and strength'', ``evidence\_span'': ``'', ``confidence'': 0.88\}}

\medskip
\noindent\textbf{Example --- unanswerable gold answer:}\\
\texttt{Question:} What is Melanie's passport number?\\
\texttt{Gold:} Not mentioned in the conversation.\\
\texttt{Output:} \texttt{\{``op'': ``NOOP''\}}
\end{promptbox}
\label{tab:prompt_repair_output}
\end{table}

\begin{table}[h]
\caption{The prompt template used for semantic consolidation (deduplication) in self-evolution.}
\begin{promptbox}[Semantic Consolidation Prompt]
\noindent\textbf{Role:} You are a memory deduplication assistant.

\medskip
\noindent\textbf{Inputs:}
\begin{itemize}[leftmargin=*]
\item A new proposed fact to be added to memory
\item One or more existing memory entries that are semantically similar (with similarity scores)
\end{itemize}

\noindent\textbf{Decision:}
\begin{itemize}[leftmargin=*]
\item \texttt{SKIP}: An existing entry already fully covers the new fact.
\item \texttt{MERGE}: The new fact describes the same event/attribute as an existing entry and adds a missing detail. Combine into one entry.
\item \texttt{INSERT}: The new fact is about a different topic, event, or time period.
\end{itemize}

\medskip
\noindent\textbf{Critical rule:} Different dates or different occurrences of the same activity $=$ \texttt{INSERT}, never \texttt{MERGE}. Only merge when both texts refer to the exact same single event at the same time.

\medskip
\noindent\textbf{Output format (JSON):}\\
\texttt{\{``action'': ``SKIP | MERGE | INSERT'', ``merge\_target\_index'': -1, ``merged\_fact'': ``'', ``reason'': ``...''\}}
\end{promptbox}
\label{tab:prompt_dedup}
\end{table}

\section{Information about AI Assistants}
We used an OpenAI LLM (GPT-5.4) as a writing and formatting assistant. In particular, it helped refine
grammar and phrasing, improve clarity, and suggest edits to figure/table captions and layout (e.g.,
column alignment, caption length, placement). The LLM did not contribute to research ideation,
experimental design, implementation, data analysis, or technical content beyond surface-level edits.
All outputs were reviewed and edited by the authors, who take full responsibility for the final text
and visuals.

\end{document}